\documentclass[journal]{IEEEtran}
\usepackage{amsmath,amssymb,amsfonts}
\usepackage{algorithmic}
\usepackage{algorithm}
\usepackage{array}
\usepackage{textcomp}
\usepackage{stfloats}
\usepackage{url}
\usepackage{verbatim}
\usepackage{graphicx}
\usepackage{cite}
\usepackage{hyperref}
\usepackage{subfigure}
\usepackage{pifont}
\usepackage[normalem]{ulem}
\useunder{\uline}{\ul}{}
\usepackage{color}
\usepackage{makecell}
\usepackage{booktabs}
\usepackage{multirow}
\usepackage{orcidlink}
\usepackage{cleveref}

\newcommand{\eg}{\textit{e.g.}}
\newcommand{\ie}{\textit{i.e.}}
\newcommand{\etal}{\textit{et al.}}

\newcolumntype{L}[1]{>{\raggedright\let\newline\\\arraybackslash\hspace{0pt}}m{#1}}
\newcolumntype{C}[1]{>{\centering\let\newline\\\arraybackslash\hspace{0pt}}m{#1}}
\DeclareMathOperator{\st}{s.t.}

\def\Deltax{\mathbf{\Delta x}}
\def\Deltaw{\mathbf{\Delta w}}
\newcommand{\cmark}{\ding{51}}
\newcommand{\xmark}{\ding{55}}

\def\rve{{\mathbf{e}}}
\def\rvx{{\mathbf{x}}}
\def\rvw{{\mathbf{w}}}
\DeclareMathOperator*{\argmax}{arg\,max}
\DeclareMathOperator*{\argmin}{arg\,min}
\def\gN{{\mathcal{N}}}

\begin{document}

\title{Improving Transferability of Adversarial Examples via Bayesian Attacks}

\author{
  Qizhang Li, Yiwen Guo, Xiaochen Yang, Wangmeng Zuo,~\IEEEmembership{Senior Member,~IEEE}, Hao Chen,~\IEEEmembership{Fellow,~IEEE}
  \IEEEcompsocitemizethanks{
  \IEEEcompsocthanksitem Q. Li is with the School of Computer Science and Technology, Harbin Institute of Technology, China. E-mail: liqizhang95@gmail.com.
  \IEEEcompsocthanksitem Y. Guo is an Independent Researcher. E-mail: guoyiwen89@gmail.com.
  \IEEEcompsocthanksitem X. Yang is with the School of Mathematics and Statistics, University of Glasgow, UK. E-mail: xiaochen.yang@glasgow.ac.uk.
  \IEEEcompsocthanksitem W. Zuo is with the School of Computer Science and Technology, Harbin Institute of Technology, China. E-mail: cswmzuo@gmail.com.
  \IEEEcompsocthanksitem H. Chen is with the Department of Computer Science, University of California, Davis, US. Email: chen@ucdavis.edu.}
\thanks{Manuscript received February 25, 2025.}
}

\markboth{Journal of \LaTeX\ Class Files,~Vol.~14, No.~8, August~2021}%
{Shell \MakeLowercase{\textit{et al.}}: A Sample Article Using IEEEtran.cls for IEEE Journals}

\IEEEpubid{\begin{minipage}{\textwidth}
  \centering
  Copyright \copyright~20xx IEEE. Personal use of this material is permitted. However, permission to use this \\ 
  material for any other purposes must be obtained from the IEEE by sending an email to pubs-permissions@ieee.org.
\end{minipage}} 


\maketitle

\begin{abstract}
The transferability of adversarial examples allows for the attack on unknown deep neural networks (DNNs), posing a serious threat to many applications and attracting great attention.
In this paper, we improve the transferability of adversarial examples by incorporating the Bayesian formulation into both the model parameters and model input, enabling their joint diversification.
We demonstrate that combination of Bayesian formulations for both the model input and model parameters yields significant improvements in transferability.
By introducing advanced approximations of the posterior distribution over the model input, adversarial transferability achieves further enhancement, surpassing all state-of-the-arts when attacking without model fine-tuning.
Additionally, we propose a principled approach to fine-tune model parameters within this Bayesian framework.
Extensive experiments demonstrate that our method achieves a new state-of-the-art in transfer-based attacks, significantly improving the average success rate on ImageNet and CIFAR-10.
Code at: \href{https://github.com/qizhangli/MoreBayesian-jrnl}{https://github.com/qizhangli/MoreBayesian-jrnl}.
\end{abstract}


\section{Introduction}
\label{sec:introduction}

\IEEEPARstart{D}{eep} neural networks (DNNs) have demonstrated remarkable performances in various applications, such as computer vision~\cite{krizhevsky2017imagenet}, natural language processing~\cite{collobert2008unified}, and speech recognition~\cite{abdel2014convolutional}. 
Nevertheless, these models have been found to be vulnerable to adversarial examples~\cite{Szegedy2014, ran2023cross, tao2024dynamics, yang2024exploring, chen2023toward, zou2023universal}, which are maliciously perturbed input data that could mislead the models into making incorrect predictions. 
This raises serious concerns about the safety of DNNs, particularly when applying them in security-critical domains. More severely, adversarial examples are transferable~\cite{Papernot2016transferability}. 
An attacker can manipulate adversarial examples on a substitute model to attack unknown victim models with different architectures and parameters. 
An in-depth exploration of adversarial transferability is crucial as it provides valuable insights into the nature of DNNs, enables more comprehensive evaluation of their robustness, and promotes the design of stronger defense methods and fully resistant networks.

Over the past several years, considerable effort has been devoted to improving the transferability of adversarial examples. In particular, ensemble-based methods~\cite{Liu2017, li2020learning,xiong2022stochastic} attract our attention as they can be readily combined with almost all other methods, such as optimizing intermediate-level representations~\cite{Huang2019,li2020yet, li2023improving, li2025enhancing} or modifying the backpropagation computation~\cite{Wu2020,guo2020back, li2024theoretical}. 
The effectiveness of ensemble-based attack is generally related to the number of models available in the bucket, thus we improve it to a statistical ensemble which consists of infinitely many substitute models in some sense by introducing a Bayesian formulation.
To model the randomness of models comprehensively, we introduce distributions to the model parameters and model input. Parameters of these distributions can be obtained from an off-the-shelf model, and if fine-tuning is possible, these parameters can further be optimized. 
Adversarial examples are then crafted by maximizing the mean prediction loss averaged across the parameters and inputs sampled from their respective distributions. 

\IEEEpubidadjcol

We evaluate our method in attacking a variety of models on ImageNet~\cite{Russakovsky2015} and CIFAR-10~\cite{Krizhevsky2009}. The proposed method achieves comparable performance without fine-tuning a substitute model, and it outperforms state-of-the-arts considerably when fine-tuning is applied. We also show that our method can be readily integrated with existing methods for further improving the attack performance.

Compared with conventional input augmentation methods~\cite{Xie2019, dong2019evading, wang2021admix, lin2019nesterov}, which apply a fixed set of transformations such as translation, scaling, or simple noise to diversify the input during the attack, our method introduces input diversity through a principled Bayesian formulation. This allows us to model input diversity via an advanced approximation of a posterior distribution learned directly from the dynamics of the iterative attack itself. 
Furthermore, this work extends parameter-only Bayesian attacks, such as our previous work~\cite{li2023making} (referred to as ``BasicBayesian'' in this paper). Specifically, \cite{li2023making} exclusively modeled the diversity of the substitute model's parameters, while the new method introduces a probability measure to model input in conjunction with the parameters of the substitute models, enabling the joint diversification of both the model and input throughout the iterative process of generating adversarial examples.
Moreover, with the incorporation of input randomness in our Bayesian formulation, we newly derive a principled optimization objective for fine-tuning that encourages flat minima in both the parameter space and the input space. 
This newly proposed optimization objective enables our method to significantly outperform state-of-the-art approaches when fine-tuning the substitute model is feasible.

\section{Background and Related Work}

\textbf{Background on adversarial examples. }
Deep neural networks are known to be vulnerable to adversarial attacks, where malicious inputs are crafted to deceive trained models. This security challenge has spurred a vast and diverse field of research. Attack paradigms, for instance, range from generating localized, perceptible patches~\cite{brown2017adversarial,ran2023cross} to crafting the subtle, full-image perturbations which are the focus of this work. The targets of these attacks are similarly varied, encompassing not only standard static models~\cite{Goodfellow2015,Szegedy2014} but also more complex, adaptive networks with dynamic architectures~\cite{tao2024dynamics}. 
Adversarial attacks have also been expanded to attack models performing various tasks, \eg, facial recognition and retrieval~\cite{tang2025towards,komkov2021advhat,dong2019efficient,li2020practical}, image compression~\cite{chen2023toward}, and person detection~\cite{xu2020adversarial}.
This ongoing threat has fueled an arms race between offense and defense, with novel attacks prompting the development of more generalizable defenses~\cite{Madry2018,zhang2019theoretically,li2022squeeze,bai2021transformers,guo2017jpeg,xie2017mitigating,fu2024remove}. To push the offensive frontier, state-of-the-art methods have explored advanced techniques, including the use of generative models to manipulate latent spaces for better transferability~\cite{gan2024lesep}. 
Among the most fundamental and widely-studied of these methods are gradient-based attacks, which iteratively optimize perturbations to maximize a model's prediction loss~\cite{li2023towards}.

Given a benign input $\rvx$ with label $y$ and a victim model $f_\rvw$ with full knowledge of architecture and parameters $\rvw$, gradient-based approaches generate adversarial examples $\rvx^{\mathrm{adv}}$ within the $\ell_p$-bounded neighborhood of $\rvx$ to maximize the prediction loss $L$:
\begin{equation}
\argmax_{\| \Deltax \|_p \le \epsilon} L(\rvx + \Deltax, y, \rvw),
\end{equation}
where $\epsilon$ is the perturbation budget.

FGSM~\cite{Goodfellow2015} is a simple one-step attack method to obtain adversarial examples in the $p=\infty$ setting: 
\begin{equation}
\rvx^\mathrm{adv} = \rvx + \epsilon\cdot\mathrm{sign}(\nabla_{\rvx}L(\rvx, y, \rvw)).
\end{equation}
The iterative variant of FGSM, I-FGSM~\cite{Kurakin2017}, is capable of generating more powerful attacks:
\begin{equation}
\rvx^\mathrm{adv}_1 = \rvx,\ \rvx^\mathrm{adv}_{t+1} = \mathrm{Clip}_\epsilon(\rvx^\mathrm{adv}_t + \eta\cdot\mathrm{sign}(\nabla_{\rvx^{\mathrm{adv}}_t}L(\rvx^{\mathrm{adv}}_t, y, \rvw)),
\end{equation}
where $\eta$ is the step size and the $\mathrm{Clip}_\epsilon$ function ensures that the generated adversarial examples remain within the pre-specified range.  

\subsection{Transfer-based Attacks} 

FGSM and I-FGSM require calculating gradients of the victim model. Nevertheless, in practical scenarios, the attacker may not have enough knowledge about the victim model for calculating gradients. To address this issue, many attacks rely on the transferability of adversarial examples, meaning that adversarial examples crafted for one classification model (using, for example, FGSM or I-FGSM) can often successfully attack other victim models.
It is normally assumed to be able to query the victim model to annotate training samples, collect a set of samples from the same distribution as that modeled by the victim models, or collect a pre-trained substitute model that is trained to accomplish the same task as the victim models.

To enhance the transferability of adversarial examples, several groups of methods have been proposed. Advancements include improved optimizer and gradient computation techniques for updating $\rvx^\mathrm{adv}$~\cite{Dong2018, lin2019nesterov, guo2020back, Huang2019, huang2022transferable, li2020yet,  guo2022intermediate, fu2024remove}, innovative strategies of training and fine-tuning substitute models~\cite{springer2021little, zhu2022toward}, and two groups of methods closely related to this paper: random input augmentation~\cite{wu2018deterministic, dong2019evading,Xie2019,lin2019nesterov,wang2021admix,wang2023structure, zhu2023boosting, wang2024boosting} and substitute model augmentation (\ie, ensemble)~\cite{Liu2017, xiong2022stochastic, gubri2022efficient, gubri2022lgv}.
For instance, in input augmentation methods, Wu \etal~\cite{wu2018understanding}, Dong \etal~\cite{dong2019evading}, Xie \etal~\cite{Xie2019}, Lin \etal~\cite{lin2019nesterov}, Wang \etal~\cite{wang2023structure}, and Wang \etal~\cite{wang2024boosting} introduced different sorts of transformations into the iterative update of adversarial examples. 
For instance, DI$^2$-FGSM~\cite{Xie2019} utilizes random resizing and padding, and TI-FGSM~\cite{dong2019evading} employs a translation-invariant methodology. Along similar lines, SI-FGSM~\cite{lin2019nesterov} enforces scale invariance, and BSR~\cite{wang2024boosting} explores robustness to structural changes through block shuffling and rotation. The effectiveness of this paradigm is further highlighted by SIA~\cite{wang2023structure}, which achieves superior results by combining multiple distinct transformation types.
GRA~\cite{zhu2023boosting} also involves input augmentation strategy where multiple inputs are sampled from the neighborhood of the original image. The final update gradient is then computed by establishing a relevance between the gradients of the original and the various augmented inputs.
Comparing with these methods, our method introduces input diversity via a principled Bayesian formulation and for the first time takes input diversity into account during substitute model fine-tuning.
For substitute model augmentation (\ie, ensemble), Liu \etal~\cite{Liu2017} proposed to generate adversarial examples on an ensemble of multiple substitute models that differ in their architectures. Additionally, Xiong \etal~\cite{xiong2022stochastic} proposed stochastic variance reduced ensemble to reduce the variance of gradients of different substitute models following the spirit of stochastic variance reduced gradient~\cite{johnson2013accelerating}. 
Gubri \etal~\cite{gubri2022lgv} suggested fine-tuning with a fixed and large learning rate to collect multiple models along the training trajectory for the ensemble attack.
In this paper, we consider the diversity in both the substitute models and model inputs by introducing a Bayesian approximation for achieving this.

\subsection{Bayesian DNNs}
\label{sec:2.2}
If a deep neural network (DNN) is considered as a probabilistic model, the process of training its parameters, denoted as $\rvw$, can be seen as maximum likelihood estimation or maximum a posteriori estimation (with regularization).
In Bayesian deep learning, the approach involves simultaneously estimating the posterior distribution of the parameters given the data.
The prediction for any new input instance is obtained by taking the expectation over this posterior distribution.
Due to the large number of parameters typically involved in DNNs, optimizing Bayesian models becomes more challenging compared to shallow models.
As a result, numerous studies have been conducted to address this issue, leading to the development of various scalable approximations.
Effective methods utilize variational inference
\cite{graves2011practical, blundell2015weight, kingma2015variational,  khan2018fast, zhang2018noisy, wu2018deterministic, xu2022variational, gao2022multi, osawa2019practical, dusenberry2020efficient}, 
dropout inference~\cite{gal2016dropout,kendall2017uncertainties, gal2017concrete},
Laplace approximation \cite{kirkpatrick2017overcoming, ritter2018scalable, li2000default},
or stochastic gradient descent~(SGD)-based approximation~\cite{mandt2017stochastic,maddox2019simple,maddox2021fast, wilson2020bayesian}. 
Taking SWAG~\cite{maddox2019simple} as an example, which is an SGD-based approximation, it approximates the posterior using a Gaussian distribution with the stochastic weight averaging~(SWA) solution as its first raw moment and the composition of a low rank matrix and a diagonal matrix as its second central moment. 
Our method is developed in a Bayesian spirit and we shall discuss SWAG thoroughly later in this paper. 
In addition to approximating the posterior over model parameters, our work also involves approximating the posterior over adversarial perturbations. This is implemented in order to introduce randomization into the model input during each iteration of iterative attacks.

In recent years, there has been research on studying the robustness of Bayesian DNNs. Besides exploring the probabilistic robustness and safety measures of such models~\cite{cardelli2019statistical, wicker2020probabilistic}, attacks have been adapted~\cite{liu2018adv, yuan2020gradient} to evaluate the robustness of these models in practice. 
While Bayesian models are often considered to be more robust~\cite{carbone2020robustness, li2019generative}, adversarial training has also been proposed for further safeguarding them, as can be seen in the work by Liu \etal~\cite{liu2018adv}. However, these studies have not specifically focused on adversarial transferability as we do in this paper.

\section{Bayesian Attack for Improved Transferability}
An intuition for improving the transferability of adversarial examples suggests improving the diversity during back-propagation. 
The BasicBayesian~\cite{li2023making} increases model diversity through Bayesian modeling, while this paper considers both model diversity and input diversity jointly.

\subsection{Generate Adversarial Examples via Bayesian Modeling}
\label{sec:3.1}

Bayesian learning aims to discover a distribution of likely models rather than a single deterministic model. 
Based on the obtained distribution of models, we can perform a transfer-based attack on an ensemble of infinitely many models.
Let $\mathcal{D} = \{(\rvx_i,y_i)\}_{i=1}^N$ denote a training set. 
Bayesian inference incorporates a prior belief about the parameters $\rvw$ through the prior distribution $p(\rvw)$ and updates this belief after observing the data $\mathcal{D}$ using Bayes' theorem, resulting in the posterior distribution: $p(\mathbf{w}|\mathcal{D})\propto p(\mathcal{D}|\rvw)p(\mathbf{w})$. 
For a new input $\rvx$, the predictive distribution of its class label $y$ is given by the Bayesian model averaging , \ie,
\begin{equation}
\label{eq:postpred}
    p(y|\rvx, \mathcal{D})=\int_{\rvw} p(y|\rvx, \rvw)p(\rvw|\mathcal{D})d\rvw,
\end{equation}
where $p(y|\rvx,\rvw)$ is the conditional probability, obtained from the DNN output followed by a softmax function. 

To perform attack on such a Bayesian model, a straightforward idea is to minimize the probability of the true class for a given input, as in~\cite{li2023making}:
\begin{equation}
\label{eq:attack}
\begin{aligned}
    &\argmin_{\|\Deltax\|_p \le \epsilon} p(y|\rvx+\Deltax, \mathcal{D}) \\ =& \argmin_{\|\Deltax\|_p \le \epsilon} \int_{\rvw} p(y|\rvx+\Deltax, \rvw)p(\rvw|\mathcal{D})d\rvw.
\end{aligned}
\end{equation}
In this paper, considering multiple-step attacks, an iterative optimizer (\eg, I-FGSM) is applied, and at the $t$-the iteration, it seeks $\Deltax_t$ to minimize
\begin{equation}
\label{eq:iterative_attack}
    \int_{\rvw} p(y|\rvx+\widehat{\Deltax}_t+\Deltax_t, \rvw)p(\rvw|\mathcal{D})d\rvw
\end{equation}
while ensuring $\|\widehat{\Deltax}_t+\Deltax_t\|_p \le \epsilon$. $\widehat{\Deltax}_1 = \mathbf{0}$, and for $t>1$, $\widehat{\Deltax}_t=\sum_{t'=1}^{t-1}\Deltax_{t'}$ is the sum of all perturbations accumulated over the previous $t-1$ iterations. 
Optimizing Eq.~(\ref{eq:attack}) or minimizing Eq.~(\ref{eq:iterative_attack}) can be regarded as generating adversarial examples that could succeed on a distribution of models, and it has been proved effective in our previous work~\cite{li2023making}.
For the rest of this paper, we shall use the notation $\rvx^{\mathrm{adv}}_t$ to represent $\rvx + \widehat{\Deltax_t}$.

In Eqs.~(\ref{eq:attack}) and~(\ref{eq:iterative_attack}), Bayesian model averaging is used to predict the label of a deterministic model input. 
Such perturbation optimization processes solely consider the diversity of models, while the diversity of model inputs is overlooked. 
Nevertheless, considering that different DNN models may be equipped with different pre-processing operations, given the same benign input (\eg, a clean image), $\rvx$ can be different for different models after their specific pre-processing steps. 
Moreover, even given the same $\rvx$ (which is the pre-processed model input), different models obtain different $\widehat{\Deltax}_t$.
Therefore, following the motivation of improving the model diversity in our previous work~\cite{li2023making}, introducing input diversity into Eqs.~(\ref{eq:attack}) and~(\ref{eq:iterative_attack}) may also be beneficial to the transferability of generated adversarial examples.

To incorporate such diversity, we simply add a randomness term $\rve$ to the input $\rvx^{\mathrm{adv}}_t$.
The randomness term should not significantly affect the model's prediction.
Therefore, the Eq.~(\ref{eq:iterative_attack}) can be rewritten as:
\begin{equation}
\label{eq:iterative_attack_random_input}
    \int_{\rvw, \rve} p(y|\rvx^{\mathrm{adv}}_t+\rve+\Deltax_t, \rvw)p(\rvw|\mathcal{D})p(\rve|\rvw,\rvx^{\mathrm{adv}}_t)d\rvw d\rve.
\end{equation}
Here the randomness term $\rve$ is added linearly to the input.
We can also introduce more complex modifications, such as by using a Gaussian filter $g(\rvx^{\mathrm{adv}}_t,\rve)$ with a random standard deviation~\cite{wang2022deepfake}; yet, for simplicity, we discuss the linear case in this paper. Due to the very large number of parameters, it is intractable to perform exact inference using Eq.~(\ref{eq:iterative_attack_random_input}). 
Instead, we adopt the Monte Carlo sampling to approximate the integral, where a set of $M$ models, each parameterized by $\rvw_j$, are sampled from the posterior $p(\rvw|\mathcal{D})$, and $S$ inputs are sampled from $p(\rve|\rvw,\rvx^{\mathrm{adv}}_t)$. 
The optimization problem can then be cast to
\begin{equation}
\begin{aligned}
\label{eq:bayesian_attack}
    \argmin_{\Deltax_t}\frac{1}{MS}\sum_{j=1}^{M}\sum_{k=1}^{S} p(y|\rvx^{\mathrm{adv}}_t+\rve_k+\Deltax_t, \rvw_j)\\
    =\argmax_{\Deltax_t}\frac{1}{MS}\sum_{j=1}^{M}\sum_{k=1}^{S} L(\rvx^{\mathrm{adv}}_t+\rve_k+\Deltax_t, y, \rvw_j), \\ \rvw_j \sim p(\rvw|\mathcal{D}), \rve_k \sim p(\rve|\rvw,\rvx^{\mathrm{adv}}_t), 
\end{aligned}
\end{equation}
where $L(\cdot, \cdot, \rvw_j)$ is a function evaluating the prediction loss of a model parameterized by $\mathbf{w}_j$.

\subsection{Construct a Posterior without Fine-tuning}
\label{sec:nofinetuning}

\begin{figure}[t]
\centering
    \includegraphics[width=0.999\linewidth]{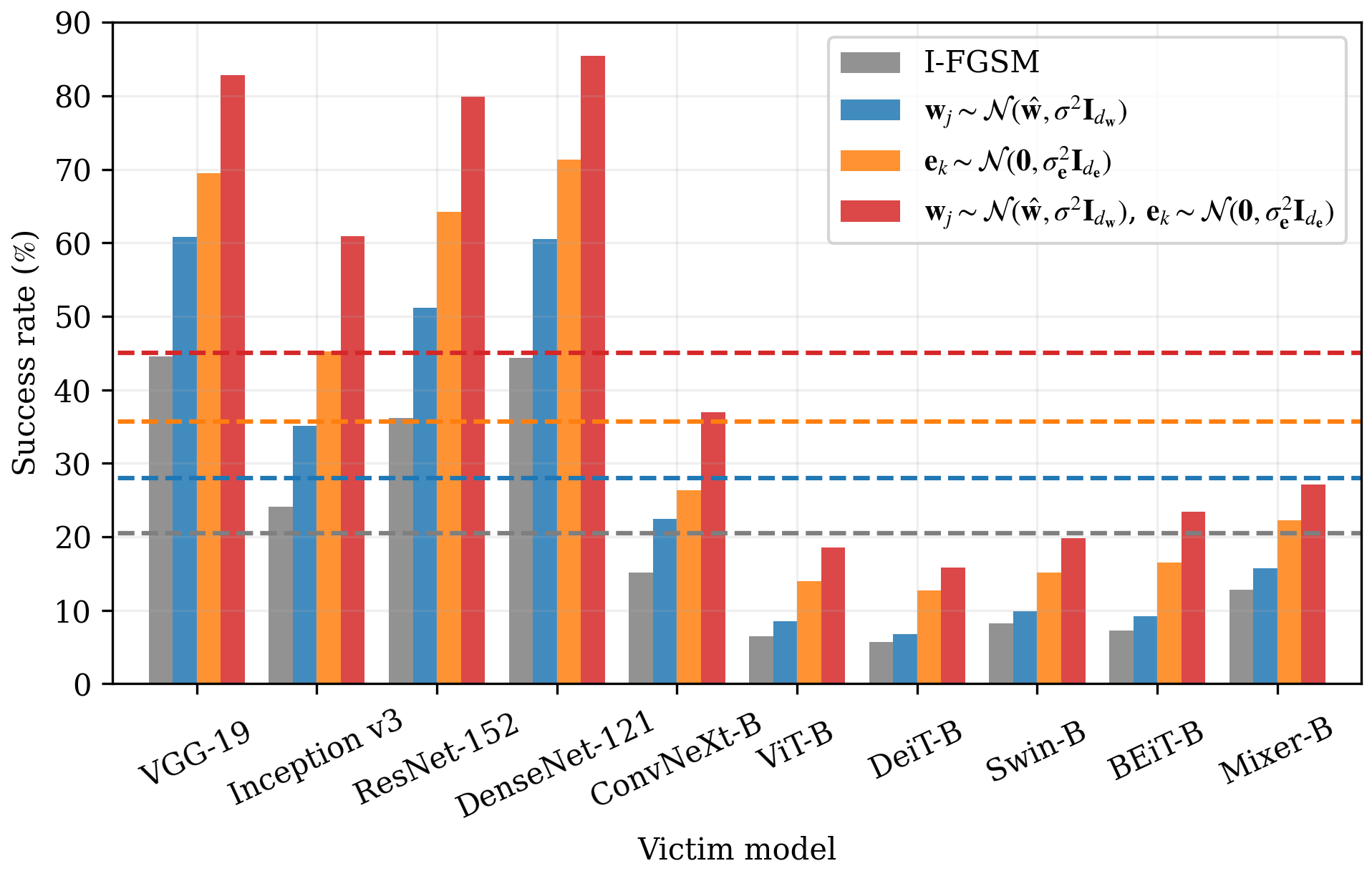}
\caption{
Comparison of success rates in attacking 10 different victim models when adversarial examples are generated on a substitute model (ResNet-50) in a non-Bayesian manner (\ie, I-FGSM) and using Bayesian modeling for model parameters $\rvw_j$, inputs $\rve_k$ (\ie, vr-IGSM~\cite{wu2018understanding}), and both $\rvw_j$ and $\rve_k$ (without fine-tuning). Dotted lines indicate the average success rate across all 10 victim models. We performed $\ell_\infty$ attacks with $\epsilon=8/255$. Best viewed in color.
}
\label{fig:1}
\end{figure}

Given any pre-trained DNN model whose parameters are $\hat{\rvw}$, we can simply obtain a posterior distribution for $\rvw_j$ without fine-tuning by assuming it follows an isotropic Gaussian~\cite{li2023making} centered at $\hat{\rvw}$, \ie, $\rvw_j \sim \mathcal{N} (\hat\rvw, \sigma^2\mathbf{I}_{d_\rvw})$, where $\sigma$ is a positive constant that controls the diversity of distribution. 
Similarly, we can consider $\rve_k \sim \mathcal{N} (\mathbf{0}, \sigma_{\rve}^2\mathbf{I}_{d_\rve})$, and it is equivalent to vr-IGSM~\cite{wu2018understanding} which suggests adding isotropic Gaussian noise into the model at each attack iteration. 
However, the improvement of joint diversification of model input and model parameters on adversarial transferability remains uncertain.
Figure~\ref{fig:1} compares the effectiveness of generating transferable attacks using I-FGSM (non-Bayesian), the original implementation of BasicBayesian~\cite{li2023making} (Bayesian over $\rvw_j$ alone), the implementation of vr-IGSM~\cite{wu2018understanding} (Bayesian over $\rve_k$ alone), and our approach in Eq.~(\ref{eq:bayesian_attack}) (Bayesian over both $\rvw_j$ and $\rve_k$); Gaussian posteriors were assumed for the latter three methods.
The experiment was conducted on ImageNet with ResNet-50 used as the substitute model; experiment details are deferred to Section~\ref{sec:setting}. 
Apparently, a joint diversification could further enhance adversarial transferability with a 17.05\% increase in the average success rate compared with considering the model diversity alone~\cite{li2023making}, and a 9.37\% increase compared with considering the input diversity alone~\cite{wu2018understanding}.
Observing the success of joint diversification, we propose a fine-tuning approach in Section~\ref{sec:finetuning} to obtain a more suitable posterior of model parameters. Additionally, we incorporate the principles of improved distribution modeling from SWAG to further improve the posteriors of both parameters and inputs in Section~\ref{sec:swag}.

\subsection{Obtain a more Suitable Posterior via Fine-tuning}
\label{sec:finetuning}

In this subsection, we explain the optimization procedure of the posterior when fine-tuning the Bayesian model from a pre-trained model is possible. Following prior work, we consider a threat model in which fine-tuning can be performed on datasets collected for the same task as the victim models.

As in Section~\ref{sec:nofinetuning}, we assume an isotropic Gaussian posterior $\mathcal{N} (\hat\rvw, \sigma^2\mathbf{I}_{d_\rvw})$; however, the mean vector $\hat\rvw$ is now considered as a trainable parameter. 
Optimization of the Bayesian model, or more specifically $\hat\rvw$, can be formulated as:
\begin{equation}
\label{eq:bayes_opt}
\max_{\hat \rvw} \frac{1}{N} \sum_{i=1}^N \mathbb{E}_{\rvw \sim \mathcal{N}(\hat\rvw, \sigma^2 \mathbf{I}_{d_\rvw}), \rve \sim \mathcal{N} (\mathbf{0}, \sigma_{\rve}^2\mathbf{I}_{d_\rve})} p(y_i|\rvx_i, \rve, \rvw).
\end{equation}
By adopting Monte Carlo sampling, it can further be reformulated as:
\begin{equation}
\begin{aligned}
\label{eq:bayes_reopt}
    \min_{\hat \rvw} \frac{1}{NMS} \sum_{i=1}^N \sum_{j=1}^M \sum_{k=1}^S L(\rvx_i+\rve_{i,k}, y_i, \hat\rvw+\Deltaw_j),\\
    \Deltaw_j \sim \gN(\mathbf{0}, \sigma^2 \mathbf{I}_{d_\rvw}), \rve_{i,k} \sim \mathcal{N} (\mathbf{0}, \sigma_{\rve}^2\mathbf{I}_{d_\rve}).
\end{aligned}
\end{equation}
The computational complexity of Eq.~(\ref{eq:bayes_reopt}) is high, thus we focus on the worst-case performance in the distributions, whose loss bounds the objective in Eq.~(\ref{eq:bayes_reopt}) from below.
The optimization problem then becomes: 
\begin{equation}
\label{eq:bayes_train}
\begin{aligned}
    & \min_{\hat \rvw} \max_{\Deltaw, \rve_i}  \frac{1}{N} \sum_{i=1}^N L(\rvx_i+\rve_i, y_i, \hat\rvw+\Deltaw), \\  & \st \, \varphi_{(\mathbf{0}, \sigma^2 \mathbf{I}_{d_\rvw})}(\Deltaw) \geq \varepsilon, \, \varphi_{(\mathbf{0}, \sigma_{\rve}^2\mathbf{I}_{d_\rve})}(\rve_i) \geq \varepsilon_{\rve}, 
\end{aligned}
\end{equation}
where $\varphi_{(\boldsymbol \mu, \boldsymbol \Sigma)}$ denotes the probability density function of a Gaussian distribution with a mean vector of $\boldsymbol \mu$ and a covariance matrix of $\boldsymbol \Sigma$; $\varepsilon$ and $\varepsilon_{\rve}$ control the confidence region of the Gaussian distributions.

To optimize the min-max problem in Eq.~(\ref{eq:bayes_train}), we first solve the inner maximization problem to find $\Deltaw^\ast$ and $\rve_i^\ast$.
Based on the first-order Taylor expansion, the loss function $L$ can be written as:
\begin{equation}
\begin{aligned}
&L(\rvx_i+\rve_i, y_i, \hat{\rvw}+\Deltaw) \\\approx &L(\rvx_i, y_i, \hat{\rvw}) + \nabla_{\hat{\rvw}} L(\rvx_i, y_i, \hat{\rvw})^\top\Deltaw + \nabla_{\rvx_i} L(\rvx_i, y_i, \hat{\rvw})^\top\rve_i.
\end{aligned}
\end{equation}
The inner maximization problem then simplifies to maximizing a linear function of $\Deltaw$ and $\rve_i$:
\begin{equation}\label{eq:max_inner_approxed}
\begin{aligned}
\max_{\Deltaw, \rve_i} \Bigg[  \frac{1}{N} \sum_{i=1}^N \left( \nabla_{\hat{\rvw}} L(\rvx_i, y_i, \hat{\rvw})\right)^\top\Deltaw \\ +  \frac{1}{N} \sum_{i=1}^N \left(\nabla_{\rvx_i} L(\rvx_i, y_i, \hat{\rvw})^\top\rve_i\right) \Bigg], \\
\st \, \varphi_{(\mathbf{0}, \sigma^2 \mathbf{I}_{d_\rvw})}(\Deltaw) \geq \varepsilon, \, \varphi_{(\mathbf{0}, \sigma_{\rve}^2\mathbf{I}_{d_\rve})}(\rve_i) \geq \varepsilon_{\rve}, 
\end{aligned}
\end{equation}
The solution to maximizing a dot product under a norm constraint is to align the vector with the gradient. 
Thus, the optimal solutions $\Deltaw^\ast$ and $\rve_i^\ast$ are in the direction of their respective gradients (\ie, $\frac{1}{N} \sum_{i=1}^N \left( \nabla_{\hat{\rvw}} L(\rvx_i, y_i, \hat{\rvw})\right)$ and $\nabla_{\rvx_i} L(\rvx_i, y_i, \hat{\rvw})$), with magnitudes as large as possible within the specified confidence regions.
These confidence regions are defined by the constraints. For example, consider the constraint $\varphi_{\mathcal{N}(\mathbf{0}, \sigma^2 \mathbf{I}_{d_\rvw})}(\mathbf{\Delta w}) \ge \epsilon$.
For an isotropic Gaussian distribution, this probability density constraint is equivalent to bounding the $\ell_2$-norm of the $\mathbf{\Delta w}$ within a hypersphere.
$\lambda_{\varepsilon, \sigma}$ is the radius of this hypersphere, \ie, $\|\mathbf{\Delta w}\|_2 \le \lambda_{\varepsilon, \sigma}$. 
It is determined by the probability $\varepsilon$ and the variance $\sigma^2$. Specifically, this radius can be formally derived by solving the constraint inequality for $\|\Deltaw\|_2$.
Starting with the constraint and the definition of the Gaussian probability density, we have
\begin{equation}
    \varphi_{\mathcal{N}(\mathbf{0}, \sigma^2 \mathbf{I}_{d_\rvw})}(\mathbf{\Delta w}) = \frac{1}{(2\pi\sigma^2)^{d_\rvw/2}} \exp\left(-\frac{\|\mathbf{\Delta w}\|_2^2}{2\sigma^2}\right) \ge \varepsilon.
\end{equation}
By taking the logarithm and rearranging the terms to solve for the norm, we find the bound for its squared value:
\begin{equation}
    \|\mathbf{\Delta w}\|_2^2 \le 2\sigma^2 \ln\left(\frac{1}{\varepsilon (2\pi\sigma^2)^{d_\rvw/2}}\right).
\end{equation}
Therefore, the radius $\lambda_{\varepsilon, \sigma}$ of the hypersphere is the square root of this bound:
\begin{equation}\label{eq:lambda}
    \lambda_{\varepsilon, \sigma} = \sqrt{2\sigma^2 \ln\left(\frac{1}{\varepsilon (2\pi\sigma^2)^{d_\rvw/2}}\right)}.
\end{equation}
The calculation of the radius $\lambda_{\varepsilon_{\rve}, \sigma_{\rve}}$ of the confidence region defined by $\varphi_{(\mathbf{0}, \sigma_{\rve}^2\mathbf{I}_{d_\rve})}(\rve_i) \geq \varepsilon_{\rve}$ proceeds analogously.
Recall that the magnitude of $\Deltaw^\ast$ and $\rve_i^\ast$ should be as large as possible within the confidence regions, we have $\|\Deltaw^\ast\|=\lambda_{\varepsilon, \sigma}$ and $\|\rve_i^\ast\|_2=\lambda_{\varepsilon_{\rve}, \sigma_{\rve}}$.
Then we obtain the approximate solution of the inner maximization in Eq.~(\ref{eq:bayes_train}):
\begin{equation}
\begin{aligned}
\Deltaw^\ast &\approx \lambda_{\varepsilon, \sigma} \frac{\frac{1}{N} \sum_{i=1}^N \nabla_{\hat{\rvw}} L(\rvx_i, y_i, \hat{\rvw})}{\|\frac{1}{N} \sum_{i=1}^N \nabla_{\hat{\rvw}} L(\rvx_i, y_i, \hat{\rvw})\|_2}, \\
\rve_i^\ast &\approx \lambda_{\varepsilon_\rve, \sigma_\rve} \frac{ \nabla_{\rvx_i} L(\rvx_i, y_i, \hat{\rvw})}{\|\nabla_{\rvx_i} L(\rvx_i, y_i, \hat{\rvw})\|_2}.
\end{aligned}
\end{equation}
Next, for the outer minimization in Eq.~(\ref{eq:bayes_train}), we have:
\begin{equation}
\begin{aligned}
\min_{\hat \rvw} \Bigg[\frac{1}{N} \sum_{i=1}^N L(x_i, y_i, \hat{\rvw}) + &\nabla_{\hat{\rvw}} L(\rvx_i, y_i, \hat{\rvw})^\top \Deltaw^\ast \\ + &\nabla_{\rvx_i} L(\rvx_i, y_i, \hat{\rvw})^\top \rve_i^\ast \Bigg].
\end{aligned}
\end{equation}
The gradient for solving this minimization problem is:
\begin{equation}
\label{eq:obj_gradient}
\nabla_{\hat\rvw} \frac{1}{N} \sum_{i=1}^N L(\rvx_i, y_i, \hat \rvw) + \mathbf{H_{\hat{\rvw}, \hat{\rvw}}}\Deltaw^\ast + \mathbf{H_{\hat{\rvw},\rvx_i}}\rve_i^\ast,
\end{equation}
which involves second-order partial derivatives in the Hessian matrix, and it can be approximately calculated using the finite difference method. 
That is, we use:
\begin{equation}
\label{eq:hessian_w}
\begin{aligned}
 \mathbf{H_{\hat{\rvw}, \hat{\rvw}}}\Deltaw^\ast \approx \frac{1}{\gamma} \left(  \nabla_{\hat\rvw} \frac{1}{N} \sum_{i=1}^N L(\rvx_i, y_i, \hat \rvw +\gamma \Deltaw^\ast) \right. \\ \left. - \nabla_{\hat\rvw} \frac{1}{N} \sum_{i=1}^N L(\rvx_i, y_i, \hat \rvw) \right),
\end{aligned}
\end{equation}
where $\gamma$ is a small positive constant. $\mathbf{H_{\hat{\rvw},\rvx_i}}\rve_i^\ast$ is approximated similarly as:
\begin{equation}
\label{eq:hessian_e}
\begin{aligned}
 \mathbf{H_{\hat{\rvw}, \rvx_i}}\rve_i^\ast \approx \frac{1}{\gamma} \left(  \nabla_{\hat\rvw} \frac{1}{N} \sum_{i=1}^N L(\rvx_i +\gamma \rve_i^\ast, y_i, \hat \rvw) \right. \\ \left. - \nabla_{\hat\rvw} \frac{1}{N} \sum_{i=1}^N L(\rvx_i, y_i, \hat \rvw) \right).
\end{aligned}
\end{equation}

By introducing $\mathbf{H_{\hat{\rvw}, \hat{\rvw}}}\Deltaw^\ast$ and $\mathbf{H_{\hat{\rvw},\rvx_i}}\rve_i^\ast$ in Eq.~(\ref{eq:obj_gradient}), we encourage flat minima in the parameter space and the input space, respectively. The former is known to be beneficial to the generalization ability of DNNs~\cite{foret2020sharpness}, while the later has been paid little attention during training/fine-tuning.

\begin{figure}[t]
\centering
    \includegraphics[width=0.999\linewidth]{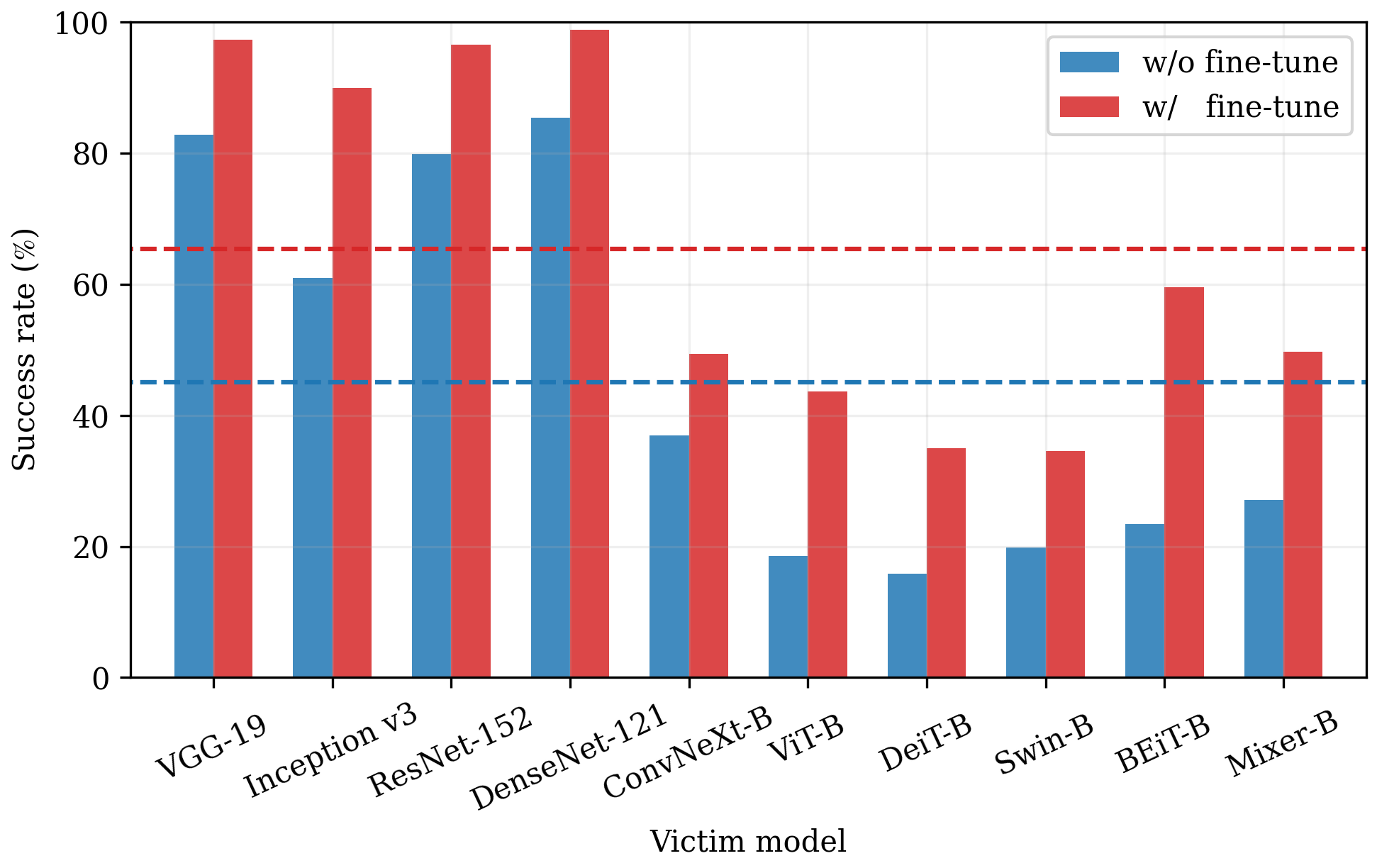}
\caption{
Comparison of adversarial attacks with and without the proposed model fine-tuning, both using isotropic Gaussian posteriors over model parameters and inputs. Dotted lines indicate the average success rates over all 10 victim models. We performed $\ell_\infty$ attacks with $\epsilon=8/255$. Best viewed in color.}
\label{fig:2}
\end{figure}

To evaluate the effectiveness of fine-tuning, an experiment was carried out on ImageNet using ResNet-50 as the substitute model, and results are shown in Figure~\ref{fig:2}. It clearly suggests that fine-tuning leads to more significant adversarial transferability. More detailed comparison results are provided in Table~\ref{tab:ft_imagenet}.

\begin{table*}[t]
\caption{Comparing transferability of FGSM and I-FGSM adversarial examples generated on a deterministic substitute model and the Bayesian substitute model (with isotropic or improved Gaussian posteriors and with or without fine-tune) under the $\ell_\infty$ constraint with $\epsilon=8/255$ on ImageNet. The architecture of the substitute models is ResNet-50, and ``Average'' was calculated over all ten victim models. The best results are marked in bold.}
\label{tab:ft_imagenet}
\begin{center}
\resizebox{0.9999\linewidth}{!}{
\renewcommand{\arraystretch}{1.}
\begin{tabular}{cccccccccccccc}
\toprule
                    &                            & \makecell{fine-\\tune}           & \makecell{VGG\\-19~\cite{Simonyan2015}}  & \makecell{Inception\\v3~\cite{Szegedy2016}} & \makecell{ResNet\\-152~\cite{He2016}} & \makecell{DenseNet\\-121~\cite{Huang2017densely}} & \makecell{ConvNeXt\\-B~\cite{liu2022convnet}} & \makecell{ViT\\-B~\cite{dosovitskiy2020image}}   & \makecell{DeiT\\-B~\cite{touvron2021training}}  & \makecell{Swin\\-B~\cite{liu2021swin}}  & \makecell{BEiT\\-B~\cite{bao2021beit}}  & \makecell{Mixer\\-B~\cite{tolstikhin2021mlp}} & {\ul Average} \\\midrule
\multirow{6}{*}{FGSM}   & -                          & \xmark                   & 41.26\% & 28.08\%      & 33.64\%    & 41.20\%      & 14.50\%    & 9.04\%  & 8.04\%  & 9.52\%  & 9.30\%  & 17.72\% & 21.23\% \\ \cmidrule{2-14}
                        & \multirow{2}{*}{Isotropic} & \xmark                   & 52.02\% & 38.10\%      & 46.40\%    & 53.84\%      & 17.06\%    & 12.50\% & 11.90\% & 13.00\% & 13.62\% & 23.18\% & 28.16\% \\
                        &                            & \cmark                   & 75.50\% & 57.14\%      & 66.08\%    & 76.78\%      & 24.94\%    & 26.02\% & 24.00\% & 23.58\% & 30.12\% & 38.04\% & 44.22\% \\\cmidrule{2-14}
                        & \multirow{2}{*}{Improved}  & \xmark & - & - & - & - & - & - & - & - & - & - & - \\
                        &                         & \cmark & \textbf{84.50\%} & \textbf{68.04\%} & \textbf{77.10\%} & \textbf{85.94\%} & \textbf{31.06\%} & \textbf{33.32\%} & \textbf{30.76\%} & \textbf{29.42\%} & \textbf{38.62\%} & \textbf{45.06\%} & \textbf{52.38\%}    \\\midrule
\multirow{6}{*}{I-FGSM} & -                          & \xmark                   & 44.52\% & 24.06\%      & 36.14\%    & 44.40\%      & 15.16\%    & 6.50\%  & 5.68\%  & 8.24\%  & 7.24\%  & 12.76\% &  20.47\% \\\cmidrule{2-14}
                        & \multirow{2}{*}{Isotropic} & \xmark                   & 82.86\% & 60.96\%      & 79.94\%    & 85.42\%      & 36.92\%    & 18.52\% & 15.86\% & 19.82\% & 23.46\% & 27.14\% &  45.09\% \\
                        &                            & \cmark                   & 97.32\% & 89.94\%      & 96.58\%    & 98.84\%      & 49.44\%    & 43.68\% & 35.00\% & 34.56\% & 59.62\% & 49.70\% &  65.47\% \\\cmidrule{2-14}
                        & \multirow{2}{*}{Improved}     & \xmark      & 92.40\% & 77.22\% & 91.42\% & 94.48\%  & 45.44\%    & 29.20\% & 26.32\% & 31.36\% & 36.74\% & 39.32\% & 56.39\% \\
                        &                            & \cmark                   &   \textbf{99.12\%} & \textbf{95.64\%} & \textbf{99.12\%} & \textbf{99.66\%} & \textbf{64.28\%} & \textbf{58.20\%} & \textbf{51.66\%} & \textbf{49.26\%} & \textbf{73.98\%} & \textbf{61.04\%} & \textbf{75.20\%}     \\\bottomrule
\end{tabular}}
\end{center} \vskip-0.05in
\end{table*} 

\subsection{Improved Distribution Modeling}
\label{sec:swag}

In Sections~\ref{sec:nofinetuning} and \ref{sec:finetuning}, we have demonstrated the superiority of adopting the Bayesian formulation for generating transferable adversarial examples. 
However, it should be noted that our approach relies on a relatively strong assumption that the posterior follows an isotropic distribution. 
Taking one step further, we remove the assumption about the covariance matrix and try to learn it from data in this subsection.

Numerous methods have been proposed to learn covariance matrices, but in this paper we opt for SWAG~\cite{maddox2019simple} due to its simplicity and scalability. 
SWAG offers an enhanced Gaussian approximation to the distribution of $\rvw$ and $\rve_i$ (for brevity, the subscript $i$ will be dropped in this subsection). 
Specifically, we adopt the SWA solution~\cite{izmailov2018} as the mean of $\rvw$, and decompose the covariance matrix into a diagonal term, a low-rank term, and a scaled identity term, \ie, $\rvw\sim \gN( \rvw_{\mathrm{SWA}}, \mathbf{\Sigma}_\rvw)$, where 
\begin{equation}
\label{eq:swag_var}
\mathbf{\Sigma}_\rvw=\alpha (\mathbf{\Sigma}_{\mathrm{diag}} + \mathbf{\Sigma}_{\mathrm{low-rank}}) +\beta \mathbf I,
\end{equation} 
$\alpha \geq 0$ represent the scaling factor of SWAG for disassociating the learning rate of the covariance~\cite{maddox2019simple}, and $\beta \geq 0$ controls the covariance matrix of the isotropic Gaussian distribution.
$\rvw_{\mathrm{SWA}}$, $\mathbf{\Sigma}_{\mathrm{diag}}$, and $\mathbf{\Sigma}_{\mathrm{low-rank}}$ are obtained after fine-tuning converges, thus we keep the fine-tuning loss and fine-tuning mechanism as in Section~\ref{sec:finetuning} even with the improved distribution modeling.  
Algorithm~\ref{alg:finetune_swag} describes the fine-tuning procedure for the substitute model, optionally incorporating SWAG.

\begin{algorithm}[t]
\caption{Fine-tuning of Substitute Model.}
\label{alg:finetune_swag}
\begin{algorithmic}[1]
\STATE \textbf{Input:} Pre-trained substitute model parameters $\hat{\rvw}$, training data $\mathcal{D}$, number of iterations $T$, SWAG start iteration $T_{\mathrm{SWA}}$, SWAG posterior update interval $c$
\STATE \textbf{Output:} Posterior distribution over model parameters
\FOR{$t=1$ \TO $T$}
    \STATE Draw a mini-batch from training set $\{\rvx_i, y_i\}_i^N \sim \mathcal{D}$
    \STATE Calculate gradient using Eqs.~(\ref{eq:obj_gradient})--(\ref{eq:hessian_e}), and update parameters $\hat{\rvw}$ with the SGD optimizer
    \STATE Set $r=t-T_{\mathrm{SWA}}$
    \IF{adopt SWAG and $r>0$ and $r \bmod c =0$}
        \STATE Update posterior over model parameters:
        \[
            p(\rvw|\mathcal{D}) \leftarrow \mathcal{N}(\rvw_{\mathrm{SWA}}, \mathbf{\Sigma}_{\rvw})
        \]
    \ENDIF
\ENDFOR
\IF{adopt SWAG}
    \STATE \textbf{return} $\mathcal{N}(\rvw_{\mathrm{SWA}}, \mathbf{\Sigma}_{\rvw})$
\ELSE
    \STATE \textbf{return} $\mathcal{N} (\hat\rvw, \sigma^2\mathbf{I}_{d_\rvw})$
\ENDIF
\end{algorithmic}
\end{algorithm}

For multi-step attacks, \eg, in I-FGSM, $\rve$ can be similarly obtained from the distribution of perturbations.
Specifically, we can obtain a Gaussian distribution of $\rvx^{\mathrm{adv}}$ using the information contained in the trajectory of a multi-step attack by the approach of SWAG, \ie, $\mathcal{N}(\rvx^{\mathrm{adv}}_{\mathrm{SWA}}, \Sigma_{\rvx^{\mathrm{adv}}})$.
The $\rvx^{\mathrm{adv}}$ from this distribution would elicit similar predictions by being fed into the model.
Therefore, for the model input at the $t$-th iteration, \ie, $\rvx^{\mathrm{adv}}_t$, we can readily obtain a posterior distribution of $\rve$ as $\mathcal{N}(\rve_{\mathrm{SWA}}, \Sigma_{\rve}) = \mathcal{N}(\rvx^{\mathrm{adv}}_{\mathrm{SWA}}-\rvx^{\mathrm{adv}}_t, \Sigma_{\rvx^{\mathrm{adv}}})$.
This posterior distribution is preferable to the isotropic distribution since it constrains the effect of the randomness term $\rve$ on the prediction.
Unlike for $\rvw_{\mathrm{SWA}}$ and $\mathbf{\Sigma}_\rvw$, it does not necessarily require model fine-tuning to obtain the mean vector and covariance matrix of the distribution, as they model the randomness in model input instead of in model parameters and the learning dynamics of the adversarial inputs are obtained during I-FGSM. 
We can start modeling the distribution and sampling from the distribution at the 10-th iteration out of 50.
Yet, for single-step attacks like FGSM, such a distribution modeling boils down to be equivalent to the isotropic case. 
Our approach to the diversity of model inputs differs from existing input augmentation techniques. Instead of simply adding noise, cropping, or translating images, we establish the posterior distribution over the input through the learning dynamics of iterative attacks and sample from this posterior distribution. Moreover, our approach can be easily combined with existing input augmentation methods to further improve performance. Experimental results are shown in Section~\ref{sec:combine}.
Algorithm~\ref{alg:attack} outlines the iterative generation of adversarial examples using the joint Bayesian framework. 

\begin{algorithm}[t]
\caption{Generation of Adversarial Examples.}
\label{alg:attack}
\begin{algorithmic}[1]
\STATE \textbf{Input:} Benign input $\rvx$ and label $y$, posterior over model parameters $p(\rvw|\mathcal{D})$, posterior over input randomness terms $p(\rve|\rvw,\rvx^{\mathrm{adv}})$, number of iterations $T$, step size $\eta$, perturbation budget $\epsilon$, number of model samples $M$, number of input samples $S$, SWAG start iteration $T_{\mathrm{SWA}}$, SWAG posterior update interval $c$
\STATE \textbf{Output:} An adversarial example 
\STATE \textbf{Initialize:} $\rvx^{\mathrm{adv}}_1 \leftarrow \rvx$, $p(\rve | \rvw, \rvx^{\mathrm{adv}}) \leftarrow \mathcal{N}(\mathbf{0}, \sigma^2_\rve \mathbf{I}_{d_{\rve}})$
\FOR{$t=1$ \TO $T$}
    \STATE Sample model parameters $\rvw_j$ and input randomness terms $\rve_k$:
    \[
    \begin{aligned}
        \mathbf{w}_j &\sim p(\mathbf{w}|\mathcal{D}) \text{\quad\quad\quad for \quad} j=1,\ldots,M \\
        \mathbf{e}_k &\sim p(\mathbf{e}|\mathbf{w},\mathbf{x}^{\mathrm{adv}}_t) \text{\quad for \quad} k=1,\ldots,S 
    \end{aligned}
    \]
    \STATE Compute input gradient of Eq.~(\ref{eq:bayesian_attack}):
    \[
    \mathbf{g}_t = \frac{1}{MS}\sum_{j=1}^{M}\sum_{k=1}^{S} \nabla_{\mathbf{x}^{\mathrm{adv}}_t} L(\mathbf{x}^{\mathrm{adv}}_t + \mathbf{e}_k, y, \mathbf{w}_j)
    \]
    \STATE Update the adversarial example:
    \[
        \mathbf{x}^{\mathrm{adv}}_{t+1} = \mathrm{Clip}_\epsilon\left(\mathbf{x}^{\mathrm{adv}}_t + \eta \cdot \mathrm{sign}(\mathbf{g}_t)\right)
    \]\vskip-0.1in
    \STATE Let $r=t-T_{\mathrm{SWA}}$
    \IF{adopt SWAG and $r > T_{\mathrm{SWA}}$ and $r \bmod c = 0$}
        \STATE Update posterior over randomness term: 
        \[
        p(\rve | \rvw, \rvx^{\mathrm{adv}}_t) \leftarrow \mathcal{N}(\rvx^{\mathrm{adv}}_{\mathrm{SWA}}-\rvx^{\mathrm{adv}}_t, \Sigma_{\rvx^{\mathrm{adv}}})
        \]
    \ENDIF
\ENDFOR
\STATE \textbf{return} $\rvx^{\mathrm{adv}}_{T+1}$
\end{algorithmic}
\end{algorithm}

We compare performance of methods with and without such improved distribution, in single-step and multi-step attacks, in Table~\ref{tab:ft_imagenet}. Approximating the covariance matrix of parameters through SWAG improves the attack success rates on all victim models, leading to an average increase of 8.16\% (from 44.22\% to 52.38\%) when using single-step attacks based on FGSM. This indicates that the more general distributional assumption of model parameters aligns better with the distribution of victim parameters in practice. A similar benefit is observed when applying the improved distribution modeling to the model input, where the average success rate increases from 45.09\% to 56.39\% in using I-FGSM.
The best performance is achieved when both improved posteriors and fine-tuning are adopted.
However, note that with fine-tuning, we suggest not to adopt such improved distribution modeling of model input and model parameters simultaneously, as the two distributions will not be independent under such circumstances and there will be more hyper-parameters to be tuned. 
Following this suggestion, in Table~\ref{tab:ft_imagenet} and subsequent experiments, when our fine-tuning, which considers both model parameters and model input, is performed, we only adopt SWAG for the model parameters and use the isotropic formulation for the model inputs during attack.

\section{Experiments}
\label{sec:experiments}
We evaluate the effectiveness of our method by comparing it to recent state-of-the-arts in this section.

\subsection{Experimental Settings}
\label{sec:setting}

To be consistent with~\cite{li2023making}, we focused on untargeted $\ell_\infty$ attacks to study the adversarial transferability.
All experiments were conducted on the same set of ImageNet~\cite{Russakovsky2015} models collected from the timm repository~\cite{rw2019timm}, \ie, ResNet-50~\cite{He2016}, VGG-19~\cite{Simonyan2015}, Inception v3~\cite{Szegedy2016}, ResNet-152~\cite{He2016}, DenseNet-121~\cite{Huang2017densely}, ConvNeXt-B~\cite{liu2022convnet}, ViT-B~\cite{dosovitskiy2020image}, DeiT-B~\cite{touvron2021training}, Swin-B~\cite{liu2021swin}, BEiT-B~\cite{bao2021beit}, and MLP-Mixer-B~\cite{tolstikhin2021mlp}. These models are well-known and encompass CNN, transformer, and MLP architectures, making the experiments more comprehensive. We randomly sampled 5000 test images that can be correctly classified by all these models from the ImageNet validation set for evaluation. Since some victim models are different from those in~\cite{li2023making}, the test images are also different. The ResNet-50 was chosen as the substitute model, same as in~\cite{li2023making}. 
For the experiments conducted on CIFAR-10, we adhered to the settings established in prior work~\cite{li2023making}. Specifically, we performed attacks on VGG-19~\cite{Simonyan2015}, WRN-28-10~\cite{Zagoruyko2016}, ResNeXt-29~\cite{Xie2017aggregated}, DenseNet-BC~\cite{Huang2017densely}, PyramidNet-272~\cite{Han2017}, and GDAS~\cite{Dong2019}, employing ResNet-18~\cite{He2016} as the substitute model. The test images used are the entire test set of the CIFAR-10 dataset.
We ran 50 iterations with a step size of 1/255 for all the iterative attacks.

In this paper, we set $\gamma$ to a fixed constant, which is slightly different from the approach described in~\cite{li2023making}. 
In \cite{li2023making}, $\gamma$ was set to be $0.1 / \| \Deltaw^* \|_2$, dependent on the value of $\| \Deltaw^* \|_2$. 
However, for our experiments on ImageNet and CIFAR-10, we set $\gamma$ to 0.1 with $\lambda_{\varepsilon, \sigma}=1$, and $\gamma$ to 0.5 with $\lambda_{\varepsilon, \sigma} = 0.2$, respectively. 
These hyper-parameters match the ones actually used in~\cite{li2023making}.
We set $\lambda_{\varepsilon_\rve, \sigma_\rve}$ to be 1 and 0.01 for ImageNet and CIFAR-10, respectively.
We used a learning rate of 0.05, an SGD optimizer with a momentum of 0.9 and a weight decay of 0.0005, a batch size of 1024 for ImageNet and 128 for CIFAR-10, and a number of epochs of 10. 
We fixed $\sigma=0.006$ and $0.012$ for ImageNet and CIFAR-10, respectively.
The $\sigma_{\rve}$ was set to be $0.01$ and $0.05$ with and without fine-tuning, respectively.
For simplicity and convenient hyper-parameter tuning, we directly set the values of $\lambda_{\varepsilon, \sigma}$ and $\lambda_{\varepsilon_\rve, \sigma_\rve}$, instead of deriving them from the confidence levels $\varepsilon$ and $\varepsilon_\rve$ according Eq.~(\ref{eq:lambda}).
Considering Eq.~(\ref{eq:swag_var}), we used $\alpha = 1$ for the posterior distribution over model parameters.
For the posterior distribution over model input, we set $\alpha=100$ and $25$ for ImageNet and CIFAR-10, respectively.
Due to the negligible difference in success rates observed between using a diagonal matrix and a combination of diagonal and low-rank matrices as the covariance for the SWAG posterior, we opted for simplicity and consistently employed the diagonal matrix. We also set $\beta = 0$ in the experiments for the same reason.
For our current method with fine-tuning, we used SWAG to approximate the posterior over model parameters, while we used an isotropic Gaussian distribution for the posterior over model inputs. 
This choice was made because modeling the posterior over model inputs becomes challenging in situations where there is a high degree of randomness in the model parameters.
We set $M=5$ and $S=5$ unless otherwise specified.

\begin{table*}[t]
\caption{
Success rates of transfer-based attacks on ImageNet using ResNet-50 as substitute architecture and I-FGSM as the back-end attack, under the $\ell_\infty$ constraint with $\epsilon=8/255$ in the untargeted setting. ``Average'' was calculated over all ten victim models. 
Detailed settings and strategies for ensuring fairness can be found in Section~\ref{sec:setting}.
The best results are marked in bold.
} \vskip-0.05in
\label{tab:comapre_imagenet}
\begin{center}
\setlength{\tabcolsep}{3pt}
\resizebox{0.999\linewidth}{!}{
\begin{tabular}{L{1.45in}C{0.45in}C{0.53in}C{0.45in}C{0.53in}C{0.53in}C{0.45in}C{0.45in}C{0.45in}C{0.45in}C{0.45in}C{0.45in}}
\toprule
\makecell[c]{Method}                    & \makecell{VGG\\-19~\cite{Simonyan2015}}  & \makecell{Inception\\v3~\cite{Szegedy2016}} & \makecell{ResNet\\-152~\cite{He2016}} & \makecell{DenseNet\\-121~\cite{Huang2017densely}} & \makecell{ConvNeXt\\-B~\cite{liu2022convnet}} & \makecell{ViT\\-B~\cite{dosovitskiy2020image}}   & \makecell{DeiT\\-B~\cite{touvron2021training}}  & \makecell{Swin\\-B~\cite{liu2021swin}}  & \makecell{BEiT\\-B~\cite{bao2021beit}}  & \makecell{Mixer\\-B~\cite{tolstikhin2021mlp}} & {\ul Average} \\\midrule
I-FGSM               & 44.52\%          & 24.06\%          & 36.14\%          & 44.40\%          & 15.16\%          & 6.50\%           & 5.68\%           & 8.24\%           & 7.24\%           & 12.76\%          & 20.47\%          \\
MI-FGSM (2018)~\cite{Dong2018}              & 54.98\%          & 34.08\%          & 46.76\%          & 55.48\%          & 20.22\%          & 10.14\%          & 8.82\%           & 11.24\%          & 10.94\%          & 17.82\%          & 27.05\%          \\
vr-IGSM (2018)~\cite{wu2018understanding} & 72.80\% & 50.02\% & 68.24\% & 75.96\% & 29.48\% & 15.18\% & 13.74\% & 16.84\% & 18.76\% & 24.44\% & 38.55\% \\
TI-FGSM (2019)~\cite{dong2019evading}              & 88.10\%          & 63.02\%          & 87.46\%          & 90.96\%          & 48.22\%          & 19.26\%          & 20.54\%          & 25.94\%          & 24.12\%          & 27.66\%          & 49.53\%          \\
DI$^2$-FGSM (2019)~\cite{Xie2019}             & 85.78\%          & 63.74\%          & 86.84\%          & 90.50\%          & 41.80\%          & 20.56\%          & 20.62\%          & 23.54\%          & 26.26\%          & 28.02\%          & 48.77\%          \\
SI-FGSM (2019)~\cite{lin2019nesterov}              & 56.32\%          & 36.40\%          & 50.34\%          & 60.70\%          & 18.58\%          & 9.16\%           & 8.18\%           & 9.86\%           & 10.00\%          & 16.64\%          & 27.62\%          \\
NI-FGSM (2019)~\cite{lin2019nesterov}              & 54.80\%          & 34.52\%          & 46.92\%          & 55.30\%          & 20.46\%          & 10.34\%          & 8.68\%           & 10.86\%          & 11.04\%          & 17.70\%          & 27.06\%          \\
ILA (2019)~\cite{Huang2019}                  & 75.30\%          & 53.54\%          & 71.40\%          & 76.88\%          & 35.34\%          & 15.02\%          & 14.04\%          & 19.10\%          & 17.34\%          & 23.00\%          & 40.10\%          \\
SGM (2020)~\cite{Wu2020}                  & 73.02\%          & 47.40\%          & 62.22\%          & 70.72\%          & 34.74\%          & 17.22\%          & 15.22\%          & 19.60\%          & 16.92\%          & 26.44\%          & 38.35\%          \\
LinBP (2020)~\cite{guo2020back}                & 77.84\%          & 51.00\%          & 63.70\%          & 75.66\%          & 24.58\%          & 10.82\%          & 8.42\%           & 12.74\%          & 13.38\%          & 20.88\%          & 35.90\%          \\
Admix (2021)~\cite{wang2021admix}                & 75.54\%          & 56.14\%          & 72.32\%          & 80.78\%          & 26.84\%          & 13.12\%          & 11.20\%          & 14.78\%          & 17.24\%          & 22.26\%          & 39.02\%          \\
NAA (2022)~\cite{zhang2022improving}                  & 79.00\%          & 63.78\%          & 72.94\%          & 82.78\%          & 27.44\%          & 13.70\%          & 12.48\%          & 16.96\%          & 17.14\%          & 26.28\%          & 41.25\%          \\
ILA++ (2022)~\cite{guo2022intermediate}                & 78.22\%          & 59.16\%          & 75.46\%          & 80.44\%          & 41.30\%          & 17.26\%          & 16.96\%          & 21.60\%          & 21.06\%          & 25.80\%          & 43.73\%          \\
PGN (2023)~\cite{ge2023boosting}               & 72.52\% & 49.36\% & 67.52\% & 75.16\% & 27.78\% & 14.94\% & 13.92\% & 15.94\% & 18.46\% & 24.50\% & 38.01\% \\
GRA (2023)~\cite{zhu2023boosting} & 75.70\%          & 55.10\%          & 72.34\%          & 78.94\%          & 31.80\%          & 17.56\%          & 17.16\%          & 19.34\%          & 21.58\%          & 27.60\%          & 41.71\%          \\
ILPD (2023)~\cite{li2023improving}                & 83.82\% & 71.26\% & 83.30\% & 87.14\% & 45.04\% & 24.30\% & 26.24\% & 30.48\% & 31.12\% & 34.60\% & 51.73\% \\
SIA (2023)~\cite{wang2023structure} & 98.12\% & 86.50\% & 97.80\% & 98.98\% & 73.60\% & 33.36\% & 32.78\% & 40.08\% & 42.50\%          & 42.24\% & 64.60\% \\
BSR (2024)~\cite{wang2024boosting} & 98.12\% & 85.92\%          & 96.42\%          & 98.60\%          & 46.86\%          & 26.06\%          & 28.86\%          & 37.34\%          & 45.34\% & 40.80\%          & 60.43\%          \\
Ours (w/o fine-tune) & 92.40\% & 77.22\% & 91.42\% & 94.48\% & 45.44\% & 29.20\% & 26.32\% & 31.36\% & 36.74\% & 39.32\% & 56.39\% \\\midrule
DRA (2022)~\cite{zhu2022toward}                  & 87.10\%          & 76.40\%          & 86.60\%          & 92.02\%          & 39.12\%          & 45.26\%          & 43.94\%          & 35.76\%          & 55.98\%          & 59.12\%          & 62.13\%          \\
LGV (2022)~\cite{gubri2022lgv}                  & 96.50\%          & 81.76\%          & 95.00\%          & 98.20\%          & 44.84\%          & 28.42\%          & 23.28\%          & 27.96\%          & 36.60\%          & 35.58\%          & 56.81\%          \\
Ours BasicBayesian (2023)~\cite{li2023making}   & 98.30\%             & 89.92\%            & 97.64\%            & 99.38\%            & 47.40\%             & 34.06\%            & 24.78\%            & 33.04\%            & 47.40\%             & 39.30\%             & 61.12\%          \\
Ours                 & \textbf{99.12\%} & \textbf{95.64\%} & \textbf{99.12\%} & \textbf{99.66\%} & \textbf{64.28\%} & \textbf{58.20\%} & \textbf{51.66\%} & \textbf{49.26\%} & \textbf{73.98\%} & \textbf{61.04\%} & \textbf{75.20\%} \\
\bottomrule
\end{tabular}}
\end{center}  \vskip-0.1in
\end{table*}

\begin{table*}[t]
\caption{
Success rates of transfer-based attacks on CIFAR-10 using ResNet-18 as substitute architecture and I-FGSM as the back-end attack, under the $\ell_\infty$ constraint with $\epsilon=4/255$ in the untargeted setting. ``Average'' was calculated over all six victim models. 
Detailed settings and strategies for ensuring fairness can be found in Section~\ref{sec:setting}.
The best results are marked in bold.
}  \vskip-0.05in
\label{tab:comapre_cifar10}
\begin{center}
\setlength{\tabcolsep}{3pt}
\resizebox{0.85\linewidth}{!}{
\begin{tabular}{L{1.45in}C{0.65in}C{0.65in}C{0.65in}C{0.65in}C{0.65in}C{0.65in}C{0.65in}}
\toprule
                 \makecell[c]{Method}              & \makecell{VGG\\-19~\cite{Simonyan2015}}           & \makecell{WRN\\-28-10~\cite{Zagoruyko2016}}              & \makecell{ResNeXt\\-29~\cite{Xie2017aggregated}}          & \makecell{DenseNet\\-BC~\cite{Huang2017densely}}         & \makecell{PyramidNet\\-272~\cite{Han2017}}       & \makecell{GDAS\\~\cite{Dong2019}}             & {\ul Average}    \\\midrule
I-FGSM                & 38.47\%          & 57.43\%          & 58.61\%          & 53.23\%          & 12.82\%          & 40.52\%          & 43.51\%          \\
MI-FGSM (2018)~\cite{Dong2018}               & 42.42\%          & 62.25\%          & 63.84\%          & 57.45\%          & 14.15\%          & 44.21\%          & 47.39\%          \\
vr-IGSM (2018)~\cite{wu2018understanding} & 59.85\% & 80.85\% & 79.84\% & 76.48\% & 20.26\% & 62.09\% & 63.23\% \\
TI-FGSM (2019)~\cite{dong2019evading}              & 50.65\%          & 74.23\%          & 74.77\%          & 69.23\%          & 18.74\%          & 53.31\%          & 56.82\%          \\
DI$^2$-FGSM (2019)~\cite{Xie2019}             & 56.99\%          & 71.50\%          & 72.90\%          & 66.51\%          & 21.57\%          & 52.22\%          & 56.95\%          \\
SI-FGSM (2019)~\cite{lin2019nesterov}              & 47.10\%          & 63.83\%          & 68.48\%          & 62.26\%          & 16.99\%          & 43.49\%          & 50.36\%          \\
NI-FGSM (2019)~\cite{lin2019nesterov}               & 42.38\%          & 62.23\%          & 63.50\%          & 57.05\%          & 14.10\%          & 44.23\%          & 47.25\%          \\
ILA (2019)~\cite{Huang2019}                  & 56.16\%          & 76.37\%          & 76.78\%          & 72.78\%          & 22.49\%          & 56.26\%          & 60.14\%          \\
SGM (2020)~\cite{Wu2020}                  & 40.85\%          & 59.61\%          & 62.50\%          & 55.48\%          & 13.84\%          & 45.17\%          & 46.24\%          \\
LinBP (2020)~\cite{guo2020back}                & 58.51\%          & 78.83\%          & 81.52\%          & 76.98\%          & 27.44\%          & 61.34\%          & 64.10\%          \\
Admix (2021)~\cite{wang2021admix}                & 56.93\%          & 75.60\%          & 78.60\%          & 73.70\%          & 21.65\%          & 52.98\%          & 59.91\%          \\
NAA (2022)~\cite{zhang2022improving}                  & 39.40\%          & 57.94\%          & 60.26\%          & 56.29\%          & 13.74\%          & 45.47\%          & 45.52\%          \\
ILA++ (2022)~\cite{guo2022intermediate}                & 60.31\%          & 77.82\%          & 78.43\%          & 74.71\%          & 26.25\%          & 59.44\%          & 62.83\%          \\
PGN (2023)~\cite{ge2023boosting}   & 58.29\% & 79.84\% & 78.52\% & 74.98\% & 19.98\% & 60.61\% & 62.04\%  \\
GRA (2023)~\cite{zhu2023boosting} & 61.30\%          & 80.56\%          & 79.20\%          & 75.86\%          & 21.15\%          & 63.06\%          & 63.52\%          \\
ILPD (2023)~\cite{li2023improving}               &       62.19\%    &     81.87\%      &       80.50\%    &      78.81\%     &      26.07\%     &     63.98\%      &       65.57\%    \\
SIA (2023)~\cite{wang2023structure} & 73.12\% & 88.27\% & 91.55\% & 86.66\% & 36.23\% & 77.83\% & 75.61\% \\
BSR (2024)~\cite{wang2024boosting} & 39.01\%          & 59.64\%          & 60.96\%          & 55.22\%          & 16.82\%          & 42.14\%          & 45.63\% \\
Ours (w/o fine-tune) & 67.44\% & 83.31\% & 84.21\% & 80.25\% & 27.83\% & 65.15\% & 68.03\% \\\midrule
LGV (2022)~\cite{gubri2022lgv}                  & 83.22\%          & 93.73\%          & 94.22\%          & 92.10\%          & 44.11\%          & 80.65\%          & 81.34\%          \\
Ours BasicBayesian (2023)~\cite{li2023making}   & 86.60\%          & 94.84\%          & \textbf{95.73\%}          & 93.78\%          & 50.11\%          & 84.97\%          & 84.34\%          \\
Ours                 & \textbf{88.04\%} & \textbf{95.00\%} & 95.62\% & \textbf{93.93\%} & \textbf{55.80\%} & \textbf{86.07\%} & \textbf{85.74\%} \\
\bottomrule
\end{tabular}}
\end{center}  \vskip-0.1in
\end{table*}

\begin{table*}[t]
\caption{
Success rates of combining our method with other state-of-the-arts on ImageNet using ResNet-50 as substitute architecture and I-FGSM as the back-end attack, under the $\ell_\infty$ constraint with $\epsilon=8/255$ in the untargeted setting. ``Average'' was calculated over all ten victim models. 
Detailed settings and strategies for ensuring fairness can be found in Section~\ref{sec:setting}.
The best results are marked in bold.
} 
\label{tab:comapre_combine}
\begin{center}
\setlength{\tabcolsep}{3pt}
\resizebox{0.999\linewidth}{!}{
\begin{tabular}{L{1.35in}C{0.45in}C{0.53in}C{0.45in}C{0.53in}C{0.53in}C{0.45in}C{0.45in}C{0.45in}C{0.45in}C{0.45in}C{0.45in}}
\toprule
\makecell[c]{Method}    & \makecell{VGG\\-19~\cite{Simonyan2015}}  & \makecell{Inception\\v3~\cite{Szegedy2016}} & \makecell{ResNet\\-152~\cite{He2016}} & \makecell{DenseNet\\-121~\cite{Huang2017densely}} & \makecell{ConvNeXt\\-B~\cite{liu2022convnet}} & \makecell{ViT\\-B~\cite{dosovitskiy2020image}}   & \makecell{DeiT\\-B~\cite{touvron2021training}}  & \makecell{Swin\\-B~\cite{liu2021swin}}  & \makecell{BEiT\\-B~\cite{bao2021beit}}  & \makecell{Mixer\\-B~\cite{tolstikhin2021mlp}} & {\ul Average} \\\midrule
Our BasicBayesian~\cite{li2023making} & 98.30\%             & 89.92\%            & 97.64\%            & 99.38\%            & 47.40\%             & 34.06\%            & 24.78\%            & 33.04\%            & 47.40\%             & 39.30\%             & 61.12\%          \\
Ours (w/o fine-tune)                    & 92.40\% & 77.22\% & 91.42\% & 94.48\%  & 45.44\%    & 29.20\% & 26.32\% & 31.36\% & 36.74\% & 39.32\% & 56.39\% \\
Ours                                    & 99.12\% & 95.64\% & 99.12\% & 99.66\% & 64.28\% & 58.20\% & 51.66\% & 49.26\% & 73.98\% & 61.04\% & 75.20\% \\\midrule

MI-FGSM (2018)~\cite{Dong2018}          & 54.98\%          & 34.08\%          & 46.76\%          & 55.48\%          & 20.22\%          & 10.14\%          & 8.82\%           & 11.24\%          & 10.94\%          & 17.82\%          & 27.05\%          \\
\quad +Our BasicBayesian~\cite{li2023making} & 98.90\% & 92.38\% & 97.92\% & \textbf{99.48\%} & 47.10\% & 39.12\% & 27.62\% & 35.26\% & 52.92\% & 45.36\% & 63.61\% \\
\quad +Ours (w/o fine-tune) & 92.58\% & 78.64\% & 91.42\% & 94.72\% & 46.38\% & 29.98\% & 26.82\% & 31.98\% & 36.98\% & 39.68\% & 56.92\% \\
\quad +Ours & \textbf{99.02\%} & \textbf{95.52\%} & \textbf{98.82\%} & \textbf{99.48\%} & \textbf{62.46\%} & \textbf{59.62\%} & \textbf{53.80\%} & \textbf{50.80\%} & \textbf{74.40\%} & \textbf{65.44\%} & \textbf{75.94\%} \\\midrule

DI$^2$-FGSM (2019)~\cite{Xie2019}                    & 85.78\%          & 63.74\%          & 86.84\%          & 90.50\%          & 41.80\%          & 20.56\%          & 20.62\%          & 23.54\%          & 26.26\%          & 28.02\%          & 48.77\%          \\
\quad +Our BasicBayesian~\cite{li2023making}         & \textbf{99.40\%} & 95.98\% & 99.36\% & \textbf{99.82\%} & 56.58\% & 53.38\% & 39.38\% & 45.96\% & 69.36\% & 54.80\% & 71.40\% \\
\quad +Ours (w/o fine-tune) & 96.62\%    & 87.70\% & 96.90\% & 98.16\%      & 55.62\%         & 40.24\%   & 38.34\% & 41.82\% & 50.40\% & 49.02\% & 65.48\% \\
\quad +Ours                                          & 99.32\%          & \textbf{96.58\%}          & \textbf{99.42\%}          & 99.70\%          & \textbf{63.40\%}          & \textbf{66.66\%} & \textbf{59.08\%} & \textbf{53.54\%} & \textbf{81.62\%} & \textbf{68.02\%} & \textbf{78.73\%} \\\midrule
SGM (2020)~\cite{Wu2020}      & 73.02\%          & 47.40\%          & 62.22\%          & 70.72\%          & 34.74\%          & 17.22\%          & 15.22\%          & 19.60\%          & 16.92\%          & 26.44\%          & 38.35\%          \\
\quad +Our BasicBayesian~\cite{li2023making} & 98.72\% & 91.30\% & 97.76\% & 99.26\% & 49.82\% & 36.10\% & 26.46\% & 34.28\% & 49.92\% & 43.06\% & 62.67\% \\
\quad +Ours (w/o fine-tune) & 94.40\% & 81.66\% & 92.34\% & 95.74\% & 50.50\% & 33.74\% & 32.30\% & 35.54\% & 41.34\% & 47.50\% & 60.51\% \\
\quad +Ours & \textbf{99.16\%} & \textbf{95.76\%} & \textbf{98.92\%} & \textbf{99.64\%} & \textbf{63.12\%} & \textbf{58.28\%} & \textbf{51.96\%} & \textbf{49.96\%} & \textbf{74.10\%} & \textbf{64.04\%} & \textbf{75.49\%} \\\midrule
LinBP (2020)~\cite{guo2020back}    & 77.84\% & 51.00\% & 63.70\% & 75.66\% & 24.58\% & 10.82\% & 8.42\% & 12.74\% & 13.38\% & 20.88\% & 35.90\%          \\
\quad +Our BasicBayesian~\cite{li2023making} & 98.66\% & 92.38\% & 97.98\% & 99.48\% & 48.62\% & 36.26\% & 24.42\% & 33.68\% & 51.02\% & 40.90\% & 62.34\% \\
\quad +Ours (w/o fine-tune) & 94.30\% & 80.94\% & 92.68\% & 95.76\% & 48.56\% & 31.58\% & 28.50\% & 34.14\% & 40.52\% & 40.58\% & 58.76\% \\
\quad +Ours & \textbf{99.16\%} & \textbf{95.72\%} & \textbf{99.14\%} & \textbf{99.68\%} & \textbf{62.94\%} & \textbf{60.84\%} & \textbf{51.74\%} & \textbf{49.88\%} & \textbf{76.10\%} & \textbf{61.78\%} & \textbf{75.70\%}  \\\midrule
ILA++ (2022)~\cite{guo2022intermediate}    & 78.22\%          & 59.16\%          & 75.46\%          & 80.44\%          & 41.30\%          & 17.26\%          & 16.96\%          & 21.60\%          & 21.06\%          & 25.80\%          & 43.73\%          \\
\quad +Our BasicBayesian~\cite{li2023making} & \textbf{99.02\%} & 93.80\% & \textbf{98.40\%} & \textbf{99.52\%} & 57.84\% & 50.58\% & 45.54\% & 50.80\% & 62.74\% & 56.30\% & 71.45\% \\
\quad +Ours (w/o fine-tune) & 92.04\%          & 80.60\%          & 91.62\%          & 93.72\%          & 48.46\% & 32.20\% & 31.38\% & 33.80\% & 41.06\% & 39.78\% & 58.47\% \\
\quad +Ours & 98.58\%          & \textbf{94.30\%} & 98.26\% & 99.18\%          & \textbf{63.06\%} & \textbf{63.28\%} & \textbf{58.56\%} & \textbf{57.12\%} & \textbf{74.96\%} & \textbf{67.40\%} & \textbf{77.47\%} \\\midrule
ILPD (2023)~\cite{li2023improving}                & 83.82\% & 71.26\% & 83.30\% & 87.14\% & 45.04\% & 24.30\% & 26.24\% & 30.48\% & 31.12\% & 34.60\% & 51.73\% \\
\quad +Our BasicBayesian~\cite{li2023making} & \textbf{99.18\%} & 96.04\% & 98.86\% & 99.54\% & 60.90\% & 54.02\% & 45.66\% & 52.50\% & 68.46\% & 59.16\% & 73.43\% \\
\quad +Ours (w/o fine-tune) & 95.58\% & 87.64\% & 95.88\% & 97.02\% & 61.00\% & 41.34\% & 40.72\% & 44.48\% & 50.34\% & 51.04\% & 66.50\% \\
\quad +Ours & 98.98\% & \textbf{96.06\%} & \textbf{98.88\%} & \textbf{99.56\%} & \textbf{66.12\%} & \textbf{67.46\%} & \textbf{62.94\%} & \textbf{60.18\%} & \textbf{80.50\%} & \textbf{70.38\%} & \textbf{80.11\%} \\\midrule
SIA (2023)~\cite{wang2023structure}    & 98.12\%          & 86.50\%          & 97.80\%          & 98.98\%          & 60.04\%          & 27.62\%          & 32.78\%          & 40.08\%          & 42.50\%          & 42.24\%          & 62.67\%          \\
\quad +Our BasicBayesian~\cite{li2023making}   & 99.72\%          & 96.58\%          & 99.64\%          & \textbf{99.90\%} & 62.46\%          & 34.32\%          & 37.78\%          & 46.70\%          & 60.40\%          & 50.54\%          & 68.80\%          \\
\quad +Ours (w/o fine-tune) & 98.56\%          & 91.68\%          & 98.36\%          & 99.22\%          & 61.92\%          & 32.56\%          & 38.90\%          & 45.66\%          & 51.98\%          & 49.80\%          & 66.86\%          \\
\quad +Ours                 & \textbf{99.76\%} & \textbf{96.88\%} & \textbf{99.78\%} & 99.82\%          & \textbf{64.54\%} & \textbf{59.56\%} & \textbf{56.18\%} & \textbf{54.36\%} & \textbf{81.46\%} & \textbf{68.96\%} & \textbf{78.13\%} \\\midrule
BSR (2024)~\cite{wang2024boosting}                   & 98.12\%          & 85.92\%          & 96.42\%          & 98.60\%          & 46.86\%          & 26.06\%          & 28.86\%          & 37.34\%          & 45.34\%          & 40.80\%          & 60.43\%          \\
\quad +Our BasicBayesian~\cite{li2023making}   & 99.34\%          & 94.50\%          & 98.42\%          & 99.62\%          & 40.30\%          & 30.96\%          & 32.20\%          & 39.76\%          & 57.18\%          & 49.12\%          & 64.14\%          \\
\quad +Ours (w/o fine-tune) & 99.72\%          & 96.98\%          & \textbf{99.32\%} & \textbf{99.84\%} & 49.66\%          & 38.58\%          & 39.08\%          & 46.42\%          & 66.98\%          & 54.64\%          & 69.12\%          \\
\quad +Ours                 & \textbf{99.78\%} & \textbf{97.40\%} & 99.22\%          & 99.66\%          & \textbf{65.40\%} & \textbf{58.98\%} & \textbf{55.40\%} & \textbf{51.84\%} & \textbf{79.84\%} & \textbf{64.66\%} & \textbf{77.22\%} \\
\bottomrule
\end{tabular}}
\end{center}  \vskip-0.1in
\end{table*}

For compared competitors, we followed their official implementations for the most of them. 
It is worth noting that recent research has found that, for input augmentation methods, aggregating more input gradients can lead to the creation of more transferable adversarial examples~\cite{zhao2022towards}. 
To ensure a fair comparison focused on the quality of different augmentation strategies, we standardized the computational budget for this class of methods by strictly controlling the number of gradient computations to be 25 per iteration.
Specifically, for our method, we used a combination of $M=5$ model samples and $S=5$ input noise samples per iteration, resulting in an average of $5 \times 5 = 25$ gradients. For methods involving random input transformations (\ie, vr-IGSM~\cite{wu2018understanding}, DI$^2$-FGSM~\cite{Xie2019}, SI-FGSM~\cite{lin2019nesterov}, Admix~\cite{wang2021admix}, PGN~\cite{ge2023boosting}, GRA~\cite{zhu2023boosting}, BSR~\cite{wang2024boosting}, and SIA~\cite{wang2023structure}),
we execute the transformation-and-gradient-computation process 25 independent times per iteration and aggregate the resulting gradients by either averaging them or integrating them with specific approaches.
For TI-FGSM~\cite{dong2019evading}, we modified its standard implementation. Instead of convolving the input gradient with a kernel, we generated 25 randomly translated input images and averaged their respective gradients, a process we found to be more effective.
In contrast, methods that do not inherently involve input augmentation (\ie, MI-FGSM~\cite{Dong2018}, NI-FGSM~\cite{lin2019nesterov}, ILA~\cite{Huang2019}, SGM~\cite{Wu2020}, LinBP~\cite{guo2020back}, NAA~\cite{zhang2022improving}, ILA++~\cite{guo2022intermediate}, and ILPD~\cite{li2023improving}) were run in their standard, single-gradient configuration. Combining them with input augmentation would turn them into a different, hybrid method, making it difficult to isolate the true benefit of their core innovations.
When comparing our method with the methods innovated by substitute model fine-tuning (\ie, LGV~\cite{gubri2022lgv}, DRA~\cite{zhu2022toward}, and our BasicBayesian~\cite{li2023making}), in order to fairly demonstrate the improvements achieved by our fine-tuning framework, for the competitors, we additionally introduced Gaussian noise into the model input multiple times and then averaged the obtained gradients. 
For LGV~\cite{gubri2022lgv} and our BasicBayesian~\cite{li2023making}, during each iteration, we sampled 5 models from the collected model set and sampled 5 Gaussian noise into the model input for each model.
For DRA~\cite{zhu2022toward}, we tested it on ImageNet using the ResNet-50 model provided by the authors, and we averaged 25 input gradients obtained by adding Gaussian noise into the model input per iteration.
We chose the best standard deviation of Gaussian noise from $\{0.01, 0.03, 0.05, 0.07, 0.09\}$ for these methods, \ie, 0.01 for LGV and our BasicBayesian~\cite{li2023making}, and 0.05 for DRA. 
By doing so, the performance of the competitors will be better compared with their default setting. 
All experiments were performed on an NVIDIA V100 GPU.

\subsection{Comparison with State-of-the-arts}
\label{sec:5.2}

We compared our method with recent state-of-the-arts in Tables~\ref{tab:comapre_imagenet} and~\ref{tab:comapre_cifar10}. 
A variety of methods were included in the comparison, including methods that adopt advanced optimizers (MI-FGSM~\cite{Dong2018} and NI-FGSM~\cite{lin2019nesterov}), increase input diversity (vr-IGSM~\cite{wu2018understanding}, TI-FGSM~\cite{dong2019evading}, DI$^2$-FGSM~\cite{Xie2019}, SI-FGSM~\cite{lin2019nesterov}, and Admix~\cite{wang2021admix}), use advanced gradient computations (ILA~\cite{Huang2019}, SGM~\cite{Wu2020}, LinBP~\cite{guo2020back}, NAA~\cite{zhang2022improving}, ILA++~\cite{guo2022intermediate}, PGN~\cite{ge2023boosting}, and ILPD~\cite{li2023improving}), and employ substitute model fine-tuning (LGV~\cite{gubri2022lgv} and DRA~\cite{zhu2022toward}). 
Detailed settings and strategies for ensuring fairness can be found in Section~\ref{sec:setting}.
The performance of all compared methods was evaluated in the task of attacking 10 victim models on ImageNet and 6 victim models on CIFAR-10.

It can be observed from the tables that the proposed method outperforms all compared methods. 
Among the methods without fine-tuning (\ie, the upper half in Table~\ref{tab:comapre_imagenet} and Table~\ref{tab:comapre_cifar10}), our method (w/o fine-tune) achieves the best average success rate of 56.39\% on ImageNet and 68.03\% on CIFAR-10.
When comparing with the substitute model fine-tuning methods (\ie, the lower half in Table~\ref{tab:comapre_imagenet} and Table~\ref{tab:comapre_cifar10}), even after combining these competitors with Gaussian noise input augmentation and controlling the number of input gradients averaged per iteration, our method still achieves the best average success rate, \ie, $75.20\%$ on ImageNet and $85.74\%$ on CIFAR-10. This shows that our fine-tuning framework can indeed boost the transferability of adversarial examples.

\subsection{Combination with Other Methods}
\label{sec:combine}
We would also like to mention that it is possible to combine our method with other attack methods to further enhance the transferability. 
In Table~\ref{tab:comapre_combine}, we report the attack success rate of our method in combination with MI-FGSM, DI$^2$-FGSM, SGM, LinBP, ILA++, and ILPD. 
Note that the number of input gradient computations is controlled to be the same in the baseline methods and after combining with our method (as detailed in Section~\ref{sec:setting}).
It can be seen that the transferability to all victim models improves substantially, compared with both the original methods and our BasicBayesian~\cite{li2023making}. 
The best performance is obtained when combining our current method with ILPD, improving over the original ILPD by $28.38\%$ and achieving a record-high success rate of $80.11\%$.

\subsection{Attacking Defensive Models}

It is also of interest to evaluate the transferability of adversarial examples to robust models, and we compared the performance of competitive methods in this setting in Table~\ref{tab:robust}. 
The victim models used in this study were collected from RobustBench~\cite{croce2020robustbench}.
All these models were trained using some sorts of advanced adversarial training~\cite{Madry2018, wong2020fast, xie2020adversarial}, and they exhibit high robust accuracy against AutoAttack~\cite{croce2020reliable} on the official ImageNet validation set. 
These models included a robust ConvNeXt-B~\cite{liu2023comprehensive}, a robust Swin-B~\cite{liu2023comprehensive}, and a robust ViT-B-CvSt~\cite{singh2023revisiting}. Following the setting in~\cite{li2023making}, two models from Bai et al.'s open-source repository~\cite{bai2021transformers}, namely a robust ResNet-50-GELU and a robust DeiT-S, were also adopted as the robust victim. The tested robust ConvNeXt-B, robust Swin-B, and robust ViT-B-CvSt show higher prediction accuracy (\ie, 55.82\%, 56.16\%, and 54.66\%, respectively, against AutoAttack) than that of the robust ResNet-50-GELU and robust DeiT-S (35.51\% and 35.50\%, respectively).

\begin{table}[t]
\caption{
Success rates of attacking adversarially trained models on ImageNet using ResNet-50 as substitute architecture and I-FGSM as the back-end attack, under the $\ell_\infty$ constraint with $\epsilon=8/255$ in the untargeted setting. ``Average'' was calculated over all five victim models.
The best results are marked in bold.
} 
\label{tab:robust}
\large
\begin{center}
\resizebox{0.99\linewidth}{!}{
\begin{tabular}{lcccccc}
\toprule
\makecell[c]{Method}        & \makecell[c]{ConvNeXt\\-B~\cite{liu2023comprehensive}}      & \makecell[c]{Swin\\-B~\cite{liu2023comprehensive}}          & \makecell[c]{ViT\\-B~\cite{singh2023revisiting}}     & \makecell[c]{ResNet\\-50~\cite{bai2021transformers}}        & \makecell[c]{DeiT\\-S~\cite{bai2021transformers}}           & \makecell[c]{{\ul Average}}          \\\midrule
Ours BasicBayesian~\cite{li2023making}   & ~6.80\%          & ~7.08\%          & ~7.22\%           & 16.86\%          & 17.64\%          & 11.12\%          \\
Ours (w/o fine-tune) & ~6.98\%          & ~7.22\%          & ~7.32\%            & 17.40\%          & 18.10\%          & 11.40\%          \\
Ours                 & \textbf{~8.14\%} & \textbf{~8.04\%} & \textbf{~8.74\%}   & \textbf{21.00\%} & \textbf{20.74\%} & \textbf{13.33\%} \\
\bottomrule
\end{tabular}}
\end{center}  \vskip-0.15in
\end{table}

We still used the ResNet-50 substitute model which was trained just as normal and not robust to adversarial examples at all.
From Table~\ref{tab:robust}, we can observe that our newly proposed method improves the transferability of adversarial examples to these defensive models, compared with the basic Bayesian formulation in~\cite{li2023making}.

Since the objective of adversarial training is different from that of normal training, in the sense that the distribution of input is different, we suggest increasing $\lambda_{\varepsilon_{\rve}, \sigma_{\rve}}$ to achieve better alignment between the distributions and to further enhance transferability. 
When we increase $\lambda_{\varepsilon_{\rve}, \sigma_{\rve}}$ from 1 to 5 or even 10, we indeed obtain even better results in attacking robust models on ImageNet, as will be discussed in the ablation study.

In addition to adversarial training, we evaluated our method against a diverse set of defenses: a certified defense, namely randomized smoothing (RS)~\cite{cohen2019certified}), which was applied to a ResNet-50 victim model, three input transformation defenses, namely Bit-Red~\cite{guo2017jpeg}, JPEG~\cite{guo2017jpeg}, and R\&P~\cite{xie2017mitigating}, all of which pre-process the adversarial input before classification, and a purification-based defense, namely NRP~\cite{naseer2020self}, which uses a dedicated model to purify the input.
In this challenging scenario, we compare our method against the high-performing competitors from our main experiments, specifically those with over a 40\% success rate shown in Table~\ref{tab:comapre_imagenet}). As shown in Table~\ref{tab:defense}, our method consistently achieves the highest success rates, confirming its superior performance against a variety of defense strategies.

\begin{table}[t]
\centering
\caption{
Attack success rates of attacking against defense methods under the $\ell_\infty$ constraint with $\epsilon=8/255$ on ImageNet in the untargeted setting. 
The best results are marked in bold.
}
\label{tab:defense}
\resizebox{0.99\linewidth}{!}{
    \renewcommand{\arraystretch}{1}
    \begin{tabular}{lccccc}
    \toprule 
Method                 & RS & Bit-Red & JPEG & R\&P & NRP \\\midrule
I-FGSM               & 12.24\%          & 15.68\%          & 18.03\%          & 20.31\%          & 13.38\%          \\
TI-IGSM~\cite{dong2019evading}              & 14.72\%          & 39.24\%          & 37.96\%          & 49.19\%          & 21.68\%          \\
DI$^2$-FGSM~\cite{Xie2019}     & 15.80\%          & 40.05\%          & 41.25\%          & 48.13\%          & 20.14\%          \\
ILA~\cite{Huang2019}                  & 13.86\%          & 32.55\%          & 32.60\%          & 40.10\%          & 16.02\%          \\
NAA~\cite{zhang2022improving}                  & 17.88\%          & 35.79\%          & 37.85\%          & 39.34\%          & 13.38\%          \\
ILA++~\cite{guo2022intermediate}                & 15.30\%          & 36.67\%          & 36.11\%          & 40.60\%          & 18.12\%          \\
GRA~\cite{zhu2023boosting}                  & 19.40\%          & 38.54\%          & 40.98\%          & 38.40\%          & 15.23\%          \\
ILPD~\cite{li2023improving}                  & 20.26\%          & 47.35\%          & 47.74\%          & 51.23\%          & 23.19\% \\
SIA~\cite{wang2023structure}                  & 17.06\%          & 55.32\% & 53.28\%          & 62.03\% & 22.40\%          \\
BSR~\cite{wang2024boosting}                  & 15.44\%          & 51.90\%          & 52.43\%          & 56.11\%          & 20.92\%          \\
Ours (w/o fine-tune) & 25.68\% & 52.85\%          & 54.80\% & 55.90\%          & 22.75\%          \\\midrule
DRA~\cite{zhu2022toward}                  & 60.94\%          & 60.24\%          & 67.42\%          & 65.70\%          & 10.58\%          \\
LGV~\cite{gubri2022lgv}                  & 22.28\%          & 52.92\%          & 55.15\%          & 58.27\%          & 16.21\%          \\
Ours-BasicBayesian~\cite{li2023making}   & 26.76\%          & 57.57\%          & 60.39\%          & 63.40\%          & 22.92\%          \\
Ours                 & \textbf{85.04\%} & \textbf{72.79\%} & \textbf{74.07\%} & \textbf{74.24\%} & \textbf{26.01\%} \\\bottomrule
    \end{tabular}
    } 
\end{table}

\begin{table}[t]
\caption{
Average success rates of attacking normally trained victim models and adversarially trained victim models on ImageNet, with different choices of $\lambda_{\varepsilon, \sigma}$ and $\lambda_{\varepsilon_{\rve}, \sigma_{\rve}}$. The substitute model is ResNet-50. The normally trained victim models are the same as those in Table~\ref{tab:comapre_imagenet}, and the adversarially trained models are the same as those in Table~\ref{tab:robust}. The symbol ``-'' signifies that the training does not converge.
The best results are marked in bold.
} 
\label{tab:ablation_lambda}
\LARGE
\begin{center}
\resizebox{0.999\linewidth}{!}{
\begin{tabular}{ccccccc}
\toprule
\multicolumn{1}{l}{}                                     &     & \multicolumn{5}{c}{{\huge $\lambda_{\varepsilon_{\rve}, \sigma_{\rve}}$}} \\
\multicolumn{1}{l}{}                                     &     & 0               & 0.5             & 1               & 5               & 10              \\
\multirow{6}{*}{{\huge $\lambda_{\varepsilon, \sigma}$}} & 0   & 47.66\%/10.86\% & 71.22\%/12.77\% & 74.33\%/14.11\% & 66.00\%/20.47\% & 47.43\%/\textbf{22.72\%} \\
                                                         & 0.1 & 50.74\%/10.93\% & 73.19\%/12.83\% & 74.08\%/14.04\% & 66.14\%/20.38\% & 48.23\%/22.70\% \\
                                                         & 0.2 & 51.97\%/11.04\% & 73.62\%/12.80\% & 74.19\%/13.84\% & 66.65\%/19.94\% & 49.57\%/22.22\% \\
                                                         & 0.5 & 54.24\%/11.14\% & 73.78\%/12.64\% & \textbf{75.60\%}/13.56\% & 67.51\%/19.09\% & 49.09\%/21.70\% \\
                                                         & 1   & 55.58\%/11.26\% & 72.81\%/12.54\% & 75.20\%/13.33\% & 65.84\%/18.48\% & -               \\
                                                         & 2   & 51.67\%/11.48\% & 66.02\%/12.66\% & 70.78\%/13.36\% & -               & -               \\
\bottomrule
\end{tabular}}
\end{center} \vskip -0.15in
\end{table}

\subsection{Ablation Study}
\label{sec:ablation}
We conduct a series of ablation experiments to study the impact of different hyper-parameters. 

\textbf{The effect of $\lambda_{\varepsilon, \sigma}$ and $\lambda_{\varepsilon_{\rve}, \sigma_{\rve}}$.}
When adopting fine-tuning, we have two main hyper-parameters that have an effect, namely $\lambda_{\varepsilon, \sigma}$ and $\lambda_{\varepsilon_{\rve}, \sigma_{\rve}}$. 
As introduced in Section~\ref{sec:finetuning}, these hyperparameters govern the breadth of the model parameter and input distributions for fine-tuning. Consequently, larger values of $\lambda_{\varepsilon, \sigma}$ and $\lambda_{\varepsilon_{\rve}, \sigma_{\rve}}$ represent training across a wider region of their respective distributions.
We conducted an empirical study to demonstrate how the performance of our method varies with the values of these two hyper-parameters on ImageNet.
We varied $\lambda_{\varepsilon, \sigma}$ from the set $\{0, 0.1, 0.2, 0.5, 1, 2\}$ and varied $\lambda_{\varepsilon_{\rve}, \sigma_{\rve}}$ from the set $\{0, 0.5, 1, 5, 10\}$ for attacking models which are the same as in Table~\ref{tab:comapre_imagenet} and Table~\ref{tab:robust}. 
The average success rates of attacking these normally trained models and robust models are given in Table~\ref{tab:ablation_lambda}.
To achieve the best performance for each $\lambda_{\varepsilon, \sigma}$ and $\lambda_{\varepsilon_{\rve}, \sigma_{\rve}}$, we tuned the other hyper-parameters using 500 randomly selected images from the validation set. These images show no overlap with the 5000 test images.

From the table, it can be observed that increasing the values of $\lambda_{\varepsilon, \sigma}$ and $\lambda_{\varepsilon_{\rve}, \sigma_{\rve}}$ from 0 to 0.5 enhances the transferability of adversarial examples in attacking normally trained models. However, excessively large values of these hyper-parameters can lead to inferior performance.
The best performance of attacking normally trained models is achieved by setting $\lambda_{\varepsilon, \sigma}=0.5$ and $\lambda_{\varepsilon_{\rve}, \sigma_{\rve}}=1$, which results in an average success rate of $75.60\%$.
However, despite achieving optimal performance, we do not adopt this setting in our main ImageNet experiments (presented in Tables~\ref{tab:ft_imagenet}, \ref{tab:comapre_imagenet}, \ref{tab:comapre_combine}, \ref{tab:robust}, and \ref{tab:defense}). Instead, we use $\lambda_{\epsilon,\sigma} = 1$ and $\lambda_{\epsilon_\rve,\sigma_\rve} = 1$ to ensure a direct and fair comparison with our prior work, BasicBayesian~\cite{li2023making}.
Considering the performance on attacking robust models, it can be observed that the average success rate peaks a larger value of $\lambda_{\varepsilon_{\rve}, \sigma_{\rve}}$.
This is partially because that reducing the prediction loss of the perturbed inputs during fine-tuning in Eq.~(\ref{eq:bayes_train}) resembles performing adversarial training, and larger $\lambda_{\varepsilon_{\rve}, \sigma_{\rve}}$ implies making the substitute model robust in a larger neighborhood of benign inputs.
A recent related method, namely DRA~\cite{zhu2022toward}, also suggests that performing regularized fine-tuning before attacking is beneficial to attacking defensive models, and it achieves an average success rate of $62.13\%$ and $17.42\%$ in attacking normally trained models and adversarially trained models, respectively. 
By setting $\lambda_{\varepsilon_{\rve}, \sigma_{\rve}} = 5$, our method has an average success rate ranging from $65.84\%$ to $67.51\%$ in attacking normally trained models and ranging from $18.48\%$ to $20.47\%$ in attacking adversarially trained models, which outperforms DRA considerably.

\begin{figure}[t]
\centering
\subfigure[w/o fine-tuning]{\includegraphics[width=0.485\linewidth]{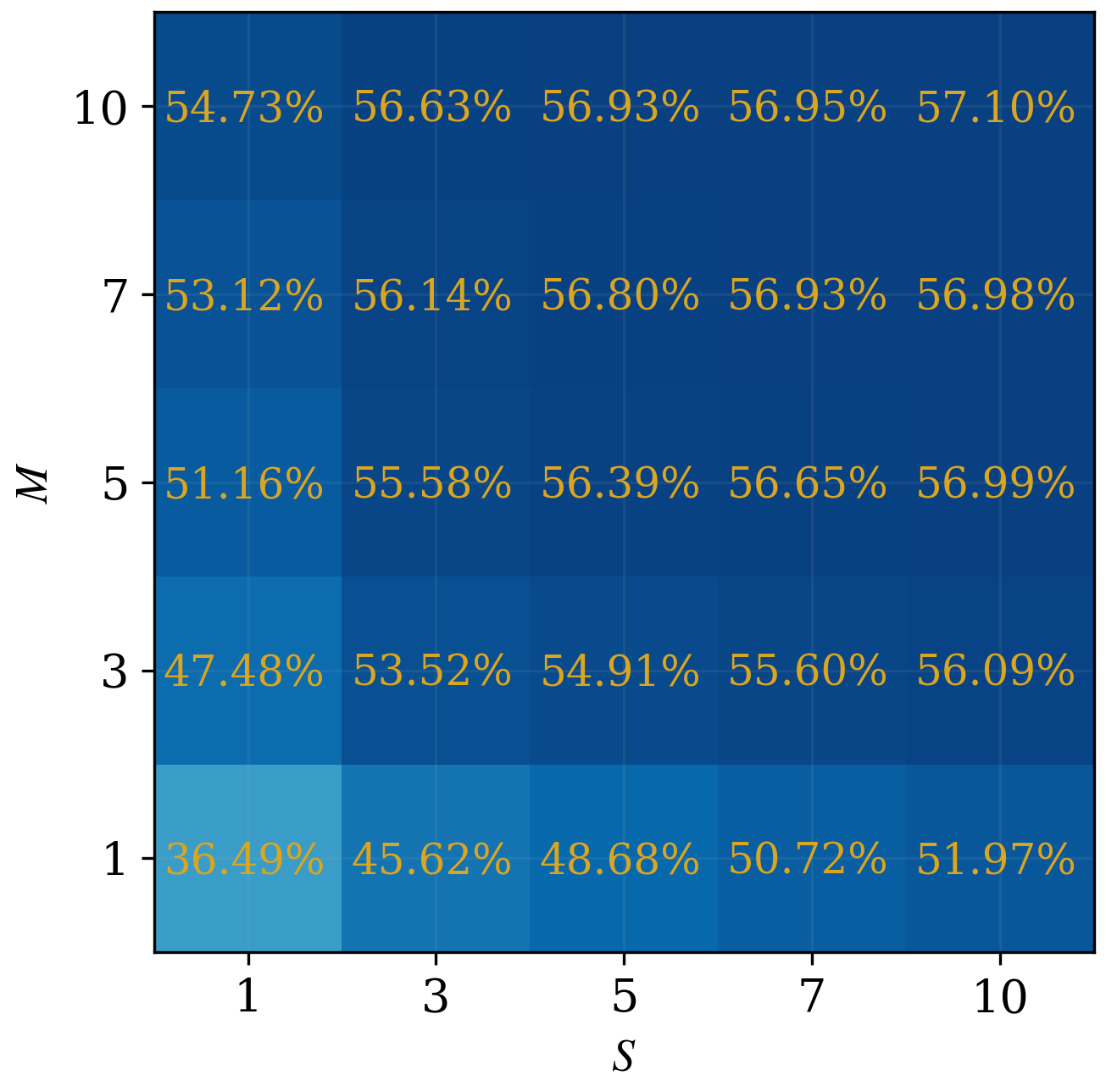}}
\subfigure[w/  fine-tuning]{\includegraphics[width=0.485\linewidth]{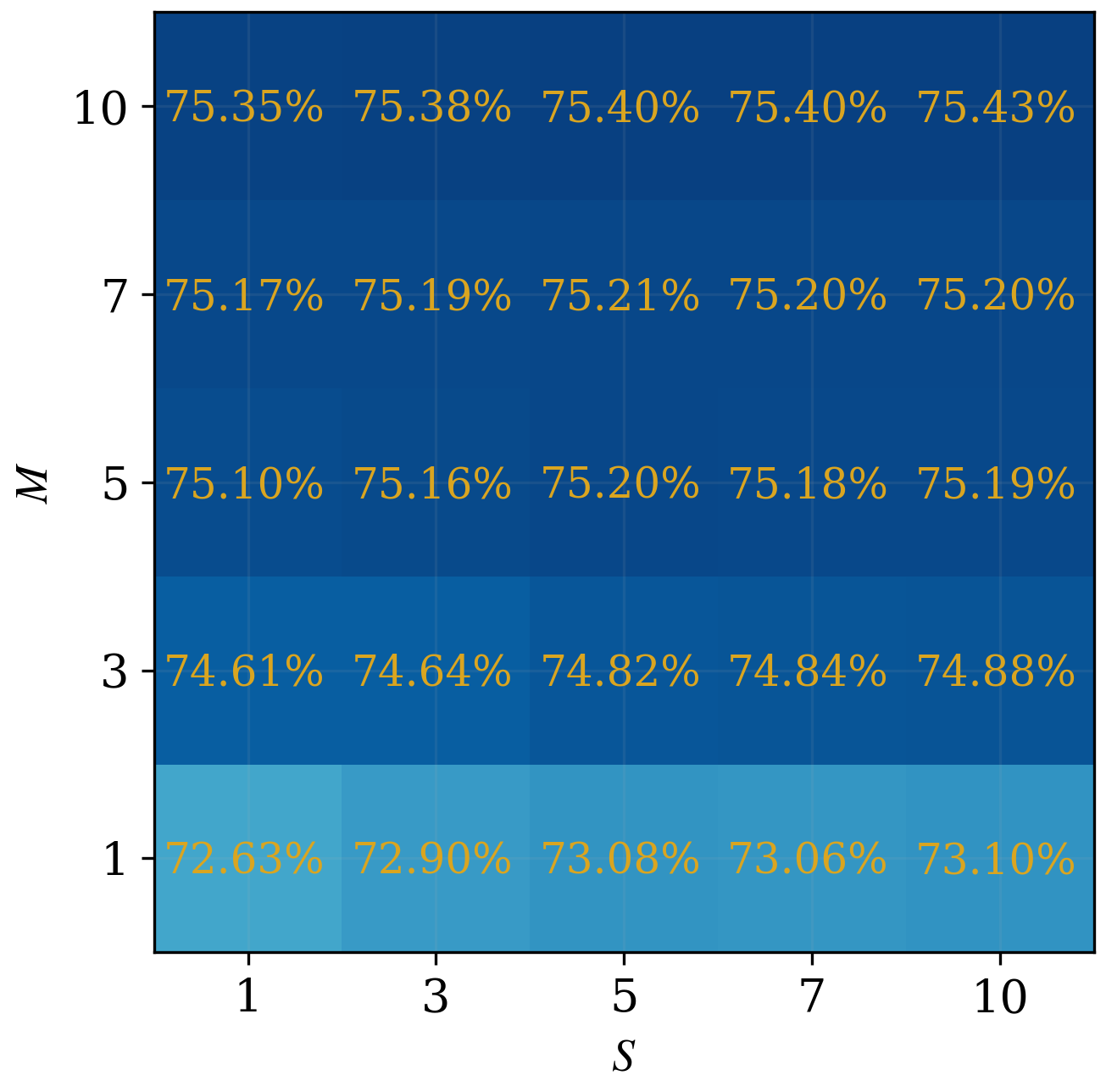}}
\caption{
Average success rates of attacking 10 victim models on ImageNet with varying $M$ and $S$. Darker cubic indicates better performance. Best viewed in color.
}\vskip-0.1in
\label{fig:ablation_MS}
\end{figure}

\textbf{The effect of $M$ and $S$.}
In Figure~\ref{fig:ablation_MS}, we conducted experiments to evaluate the transferability of adversarial examples crafted using different choices of the number of model parameters and input noise sampled at each attack iteration, \ie, $M$ and $S$, in the cases of with and without fine-tuning.
The average success rates were obtained by attacking the same victim models as in Table~\ref{tab:comapre_imagenet}.
The results apparently show that sampling more substitute models and more input noise can indeed enhance the transferability of adversarial examples, just as expected. Note that our fine-tuning framework which considers the joint diversification of both the model input and model parameters can outperform our previous work~\cite{li2023making} even when setting $M=1$ and $S=1$. Our method with this configuration achieves a higher attack success rate (72.63\%) compared to the previous approach that used 5 model parameters and 5 input noise samples at each iteration (61.12\%), indicating that our method can generate more transferable adversarial examples with lower computational complexity compared to our previous work~\cite{li2023making}.

\section{Conclusion}
In this paper, we aim at improving the transferability of adversarial examples. We have developed a Bayesian formulation for performing attacks, which can be equivalently regarded as generating adversarial examples on a set of infinitely many substitute models with input augmentations. We also advocated possible fine-tuning and advanced posterior approximations for improving the Bayesian model. 
Extensive experiments have been conducted on ImageNet and CIFAR-10 to demonstrate the effectiveness of the proposed method in generating transferable adversarial examples. 
It has been shown that our method outperforms recent state-of-the-arts by large margins in attacking more than 10 DNNs, including convolutional networks and vision transformers and MLPs, as well as in attacking defensive models.
We have also showcased the compatibility of our method with existing transfer-based attack methods, leading to even more powerful adversarial transferability. 

\bibliographystyle{IEEEtran}
\bibliography{ref}

\vfill

\end{document}